\documentclass[journal]{IEEEtran}

\usepackage{graphicx}
\usepackage{amsmath,amssymb} 
\usepackage{url}

\usepackage{breakurl}
\usepackage[breaklinks]{hyperref}
\usepackage{array}
\usepackage{float}
\usepackage{booktabs}
\usepackage{comment}
\usepackage{subcaption}
\usepackage{color}
\usepackage{flushend}
\usepackage{epstopdf}
\usepackage{multirow}
\usepackage{bm}
\usepackage{paralist}
\usepackage{color}
\usepackage{hyperref}

\newcommand{\Outer}{Element-wise}
\newcommand{\EleAttGn}{{EleAttG}}
\newcommand{\EleAttG}{{EleAttG~}}
\newcommand{\EARNN}{EleAtt-RNN}

\ifCLASSINFOpdf
\else
\fi

\begin{document}
	
%
\title{EleAtt-RNN: Adding Attentiveness to Neurons in Recurrent Neural Networks}
%
%
%

\author{{Pengfei Zhang},
	Jianru Xue,~\IEEEmembership{Member,~IEEE,}
	Cuiling Lan,~\IEEEmembership{Member,~IEEE,}
	Wenjun Zeng,~\IEEEmembership{Fellow,~IEEE,}
	{Zhanning Gao},
	Nanning Zheng,~\IEEEmembership{Fellow,~IEEE}
	
\thanks{P. Zhang, J. Xue, and N. Zheng are with Xian Jiaotong University, Xian, Shannxi, P. R. China. E-mail: zpengfei@stu.xjtu.edu.cn, \{jrxue,nnzheng\}@mail.xjtu.edu.cn}
\thanks{C. Lan, W. Zeng are with Microsoft Research Asia, Beijing, P. R. China. \protect E-mail: \{culan,wezeng\}@microsoft.com}
\thanks{Z. Gao is with Alibaba Group, Beijing, P.R. China. E-mail: zhanninggao@gmail.com}
\thanks{Corresponding authors: Cuiling Lan and Jianru Xue}}

%
%

\markboth{IEEE TRANSACTIONS ON IMAGE PROCESSING,~Vol.~XX, No.~XX, XX~XXX}%
{Shell \MakeLowercase{\textit{et al.}}: Bare Demo of IEEEtran.cls for IEEE Journals}
%



\maketitle

\begin{abstract}
Recurrent neural networks (RNNs) are capable of modeling temporal dependencies of complex sequential data. In general, current available structures of RNNs tend to concentrate on controlling the contributions of current and previous information. However, the exploration of different importance levels of different elements within an input vector is always ignored. We propose a simple yet effective Element-wise-Attention Gate (EleAttG), which can be easily added to an RNN block (e.g. all RNN neurons in an RNN layer), to empower the RNN neurons to have attentiveness capability. For an RNN block, an \EleAttG is used for adaptively modulating the input by assigning different levels of importance, {\it i.e.}, attention, to each element/dimension of the input. We refer to an RNN block equipped with an \EleAttG as an EleAtt-RNN block. Instead of modulating the input as a whole, the \EleAttG modulates the input at fine granularity, {\it{i.e.}}, element-wise, and the modulation is content adaptive. The proposed \EleAttGn, as an additional fundamental unit, is general and can be applied to any RNN structures, {\it e.g.}, standard RNN, Long Short-Term Memory (LSTM), or Gated Recurrent Unit (GRU). We demonstrate the effectiveness of the proposed EleAtt-RNN by applying it to different tasks including the action recognition, from both skeleton-based data and RGB videos, gesture recognition, and sequential MNIST classification. Experiments show that adding attentiveness through EleAttGs to RNN blocks significantly improves the power of RNNs.	
\end{abstract}

\IEEEpeerreviewmaketitle
\section{Introduction}

\IEEEPARstart{E}{xploration} of the temporal dynamics and spatial correlations of time series data plays an important role in sequence understanding. Recurrent neural networks, based on recursive connection, are powerful in modeling temporal dynamics and learning appropriate feature representations. Standard RNN (sRNN) and its variants, including Long Short-Term Memory (LSTM) \cite{hochreiter1997long}, Gated Recurrent Unit (GRU) \cite{cho14}, etc., have proven effective for tasks using sequential information, such as action recognition \cite{du2015hierarchical}, machine translation \cite{cho14},  pedestrian trajectory prediction \cite{zhang2019sr}, video summation \cite{zhao2018hsa}, and image caption \cite{vinyals2015show}.

\begin{figure*}[!t]
	\centering
	\begin{subfigure}[t]{0.4\linewidth}
		\centering\includegraphics[width=\textwidth]{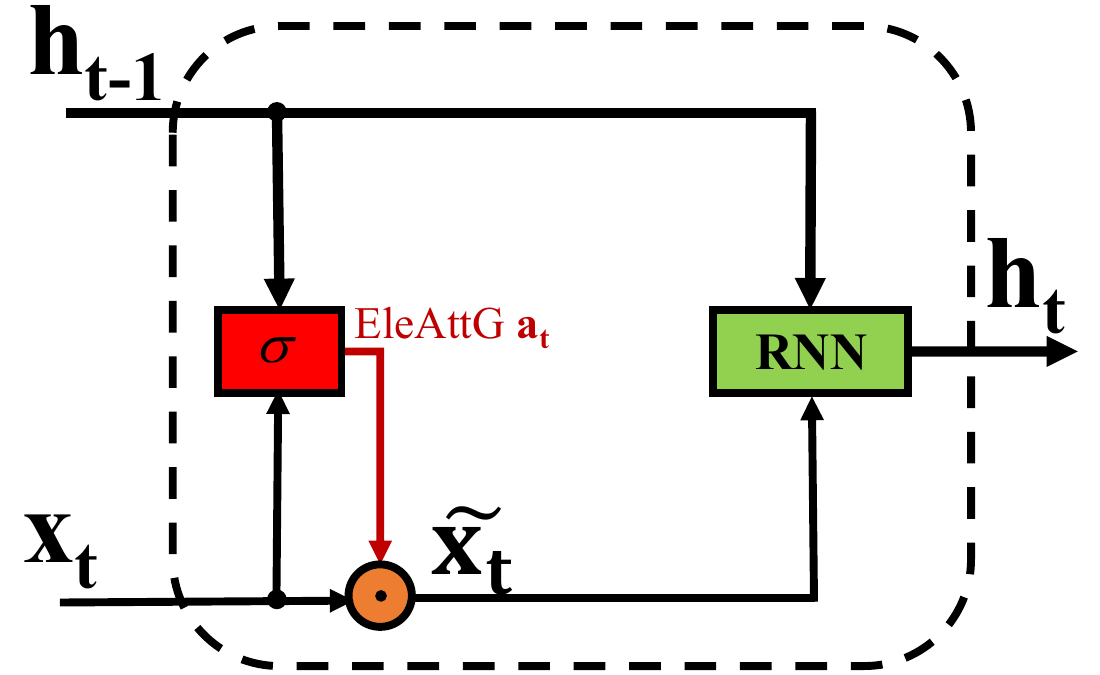}
		\caption{}
		\label{subfig:generalRNN}
	\end{subfigure}	
	\hfil
	\begin{subfigure}[t]{0.45\linewidth}
		\centering\includegraphics[width=\textwidth]{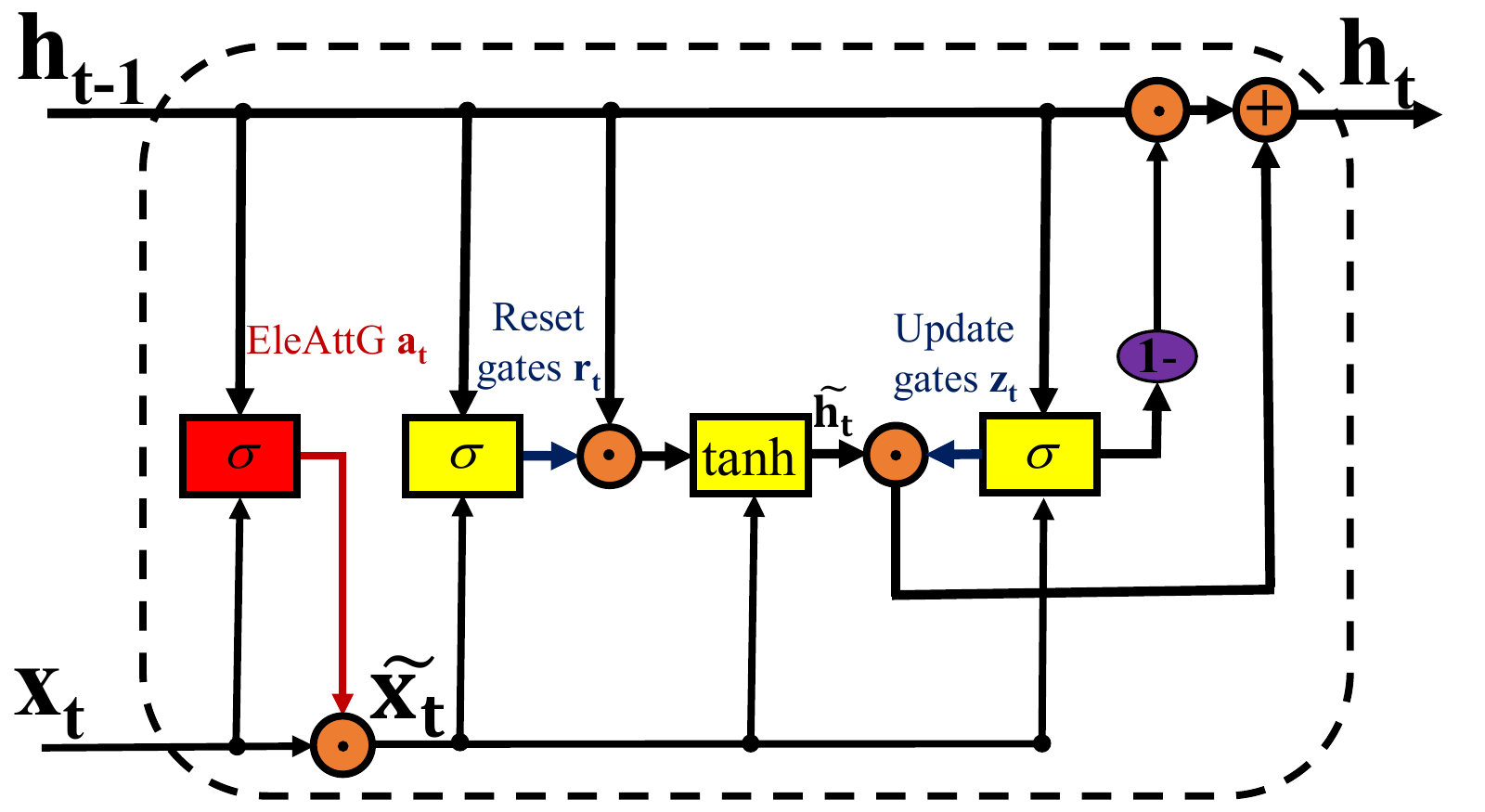}
		\caption{}			
		\label{subfig:EleGGRU}
	\end{subfigure}
	\caption[]{Illustration of \Outer-Attention Gate (EleAttG) (marked in red) for (a) a generic RNN block, where the RNN structure could be the standard RNN, LSTM, or GRU and (b) a GRU block which consists of a group of ({\it e.g.,} $N$) GRU neurons. In the diagram, each line carries a vector. The brown circles denote element-wise operation, {\it e.g.,} element-wise vector product or vector addition. The yellow boxes denote the units of the original GRU with the output dimension of $N$. The red box denotes the EleAttG with an output dimension of $D$, which is the same as the dimension of the input $\bf{x}_t$.}\label{fig:IAttRNN}
\end{figure*}

Recursive connection inside RNN structures is achieved by taking the output of the previous time step as the input of the current time step, which facilitates the processing of sequential data. At each time step, RNN neurons perform the same operation to embed the current input and the historical information to the representation of the hidden state of this time step. However, the sRNN suffers from gradient vanishing, which has difficulties in learning long-range dependencies. To address this problem, researchers propose some gate-based RNN structures, such as LSTM and GRU, which introduce gates and linear memory units inside RNN neurons to control the information flow. Gates provide a way to optionally let information through or stop softly, which balances the contributions of the information of the current time slot and historical information. For an RNN neuron, a gate produces a scalar value ranging from 0 to 1 which controls the amount of information flow. However, it is a scalar which imposes the same control on each element of a vector, rather than element-wise adaptation. They are not capable of exploring the potential different characteristics of different elements.

Attention mechanisms which selectively focus on different parts of the data have proven effective for many tasks \cite{luong2015effective,vaswani2017attention,xu2015show,li2017attentive,sharma2015actionattention,wang2016hierarchical}. Inspired by both the attention and gate mechanisms, we develop an \Outer-Attention Gate (\EleAttGn) to empower the RNN neurons to have the attentiveness capability, which makes the RNN neurons focus on the important elements of inputs adaptively. For all neurons of an RNN block, we design a sharable {\EleAttGn} which outputs an attention vector with the same dimension as the input. Then the original input is modulated by the attention vector to strengthen the impact of important elements while suppressing the impact of unimportant elements. Note that similar to \cite{LSTMblog}, we use an RNN block to represent an ensemble of a group of $N$ RNN neurons, which could be all RNN neurons in an RNN layer.

Fig. \ref{fig:IAttRNN} illustrates the structure of the proposed {\EleAttGn} within a general RNN block, as shown in (a), and a special case when the RNN block is constrained as GRU, as shown in (b). The response $\bf{a}_t$ of the {\EleAttGn} is used to modulate the input $\bf{x}_t$ to $\widetilde{\bf{x}_t}$. Then $\widetilde{\bf{x}_t}$ takes the place of the role of $\bf{x}_t$ for subsequent operations. We refer to an RNN block equipped with an EleAttG as an EleAtt-RNN block. According to the RNN structure adopted in the EleAtt-RNN block, {\it e.g.}, standard RNN, LSTM, and GRU, we refer to the blocks as EleAtt-sRNN, EleAtt-LSTM, and EleAtt-GRU, respectively. An RNN layer with such EleAttG can replace the original RNN layer and multiple EleAtt-RNN layers can be stacked.

We demonstrate the effectiveness and good generalization of our proposed EleAtt-RNN by applying it to three different tasks including action recognition, gesture recognition, and sequential MNIST classification. For action recognition, we evaluate EleAttG on both 3D skeleton-based human action recognition and RGB-based action recognition tasks. For 3D skeleton-based human action recognition, we build the classification network by stacking several EleAtt-RNN layers and evaluate the effectiveness of {\EleAttG} on three types of RNN structures, {\it{i.e.}}, standard RNN, LSTM, and GRU, respectively. EleAtt-RNNs consistently outperform the original RNNs for all the three types of RNNs. Our scheme based on EleAtt-GRU achieves state-of-the-art performance on three challenging datasets, {\it i.e.,} the NTU~\cite{Shahroudy_2016_CVPR}, N-UCLA~\cite{wang2014cross}, and SYSU~\cite{hu2015jointly} datasets. For RGB-based action recognition, we design our system by applying an EleAtt-GRU network to the sequence of frame-level CNN features. Experiments on both the JHMDB \cite{jhuang2013towards} and NTU \cite{Shahroudy_2016_CVPR} datasets show that adding {\EleAttGn}s brings significant gain. Experiments on the gesture recognition dataset DHG \cite{de2016skeleton} prove {\EleAttG} is also helpful for gesture recognition. We also evaluate the effectiveness of the EleAttG on sequential MNIST classification, which is widely used to evaluate the performance of different RNN structures \cite{arjovsky2016unitary,cooijmans2016recurrent,campos2018skip,li2018independently}. 

We summarize the merits of the proposed EleAttG as follows:
\begin{itemize}
    \setlength{\itemsep}{0pt}
    \item  \EleAttG is capable of adaptively modulating the input at a fine granularity, paying different levels of attention to different elements of the input, resulting in higher performance.
    \setlength{\itemsep}{0pt}
    \item  The design is very simple. For an RNN layer, only one line of code needs to be added in the implementation. 
    \setlength{\itemsep}{0pt}
    \item The EleAttG is general and can be added to any RNN structure, {\it{e.g.}}, standard RNN, LSTM, and GRU, and to any layer. EleAttG can be applied to multiple tasks and different input types.
    \setlength{\itemsep}{0pt}
\end{itemize}

It should be noted that this paper is an extension of our previous conference paper \cite{zhang2018adding}. In our conference paper, we only apply the proposed EleAttG to the action recognition based on skeleton data and CNN features of RGB videos. As an extension, we evaluate the generality of EleAttG by applying it to two additional tasks including gesture recognition and sequential MNIST classification. Sequention MNIST classification is widely used to evaluate the performance of RNNs. The superior performance on different tasks and datasets demonstrates the effectiveness and good generalization performance of the ``EleAtt-RNN". We also add visualization analyses and training loss curve on the sequential MNIST dataset  to help understand the effectiveness of EleAttG on different types of input data.  We also discuss the influence of EleAttG on the input vector and the hidden state, respectively. In addition, we discuss the difference between the EleAttG and the input gate of LSTM to have a better understanding of our proposed EleAttG.   

\section{Related work}
\subsection{Recurrent Neural Networks}

Recurrent neural networks have many variants. To address the vanishing gradient problem that exists in the standard RNN, Hochreiter {\it et al.} propose LSTM, which adds a memory cell that allows ``constant error carrousels" and introduces several gates to open and close access to the constant error flow \cite{hochreiter1997long}. Gers {\it et al.} propose the ``forget gate" for the previous LSTM to enable the LSTM cell to learn to reset itself (historical information) to prohibit the growth of the state indefinitely \cite{gers1999learning}. Another variant of LSTM is the peephole LSTM, which lets the gates look at the cell state \cite{gers2002learning}. GRU, is a much simpler variant of LSTM. A GRU has a reset gate and an update gate which control the memory and new input information. Comparing LSTM with GRU, there is no clear winner \cite{chung2014empirical,jozefowicz2015empirical} which differs for different application tasks. A differential gating scheme is introduced in LSTM in \cite{veeriah2015differential} which leverages the derivative of the cell state to gate the information flow, and proves effective for action recognition. Residual connection is another effective way to solve the vanishing gradient problem. Campos {\it{et al.}} propose the Skip RNN with residual connection to skip some state updates during both training and test procedures, which shortens the effective size of the computation graph \cite{campos2018skip}. Kusupati {\it{et al.}} add the residual connection to connect the current hidden state and previous hidden state \cite{kusupati2018fastgrnn}. In addition, they share the same matrices for computing the hidden state and gate value to reduce model size.

In this work, we augment the capability of RNNs by adding attentiveness to the RNN structures. We propose a simple yet effective {\EleAttGn} which adaptively modulates the elements of inputs in fine-grained manner to explore their different blue levels of importance for an RNN block.

\subsection{Attention Mechanisms}
Attention mechanisms which selectively focus on different parts of the data have proven effective for many tasks such as machine translation \cite{luong2015effective,vaswani2017attention}, image caption \cite{xu2015show}, object detection \cite{li2017attentive}, and action recognition \cite{sharma2015actionattention,wang2016hierarchical}.

For machine translation, Luong {\it et al.} study some simple attention methods. The attention model infers the attention weights at each step, and utilizes the weights to average the embedding vectors of the source words \cite{luong2015effective}. For image caption, Xu {\it et al.} split an image into several parts with each part represented by the CNN features. To enable the decoder focus more on the informative parts of the image, at each time step, the attention module outputs attention weights with respect to those parts and the weighted average is taken as the input to the decoder at that time step \cite{xu2015show}. A similar idea is adopted in RGB-based action recognition in \cite{sharma2015actionattention}. The above attention models mainly focus on how to  average a set of feature vectors with suitable weights to generate a pooled vector of the same dimension as the input to RNN. However, they do not consider the fine-grained adjustment based on different levels of importance across the input dimensions. In addition, they address attention at the network level, but not RNN block level.

For skeleton-based action recognition,  Weng {\it et al.} propose a ST-NBNN model to identify the key temporal stages and spatial joints with bilinear classifier \cite{weng2017spatio}. Liu {\it et al.} propose a global context-aware attention module to assign different joints with different attention weights \cite{liu2017global}. Since the global information of a sequence is required to learn the attention weights, the system suffers from time delay. Song {\it et al.} propose a spatio-temporal attention model without requiring global information \cite{song2017end}. A spatial attention subnetwork is used to modulate the skeleton input to selectively focus on discriminative joints before being fed into the main classification network. However, their designs of \cite{liu2017global, song2017end}are not general and have not been extended to higher RNN layers. In contrast, our proposed enhanced RNN, with EleAttG included as a fundamental unit of RNN block, is general, simple yet effective, which can be applied to any RNN block/layer. 

Note that SENet \cite{hu2018squeeze} and CBAM \cite{woo2018cbam} share similar high level ideas with our proposed EleAttG. Hu {\it et al.} design a squeeze-and-excitation attention module to explore the interdependencies of channels  by enhancing the influence of important channels and suppressing the influence of trivial channels. Woo {\it et al.} decouple the attention module by performing spatial-wise attention and channel-wise attention separately. SENet and CBAM are designed to selectively focus on important channels or/and spatial positions for CNN-based networks. The investigations on efficient attention designs for RNNs are still under-exploited. We propose a simple yet effective attention gate ({\it i.e.,} EleAttG) for RNNs, e.g., standard RNN, LSTM, and GRU.

\subsection {Action Recognition and Gesture Recognition}
\label{subsec:action}
Many traditional approaches focus on how to design efficient features to solve the problems of small inter-class variation, large view variations, and the modeling of complicated spatial and temporal evolution  \cite{wang2013action,wang2014cross,vemulapalli2014human,xia2012view,wang2016graph, de2016skeleton}. 

For skeleton-based action recognition and gesture recognition, Du {\it et al.} propose a hierarchical RNN model with the hierarchical body partitions as inputs to different RNN layers \cite{du2015hierarchical}. In order to exploit the co-occurrence of discriminative joints, Zhu {\it et al.} propose a deep regularized LSTM networks with group sparse regularization \cite{zhu2016co}. In addition, they add inner dropout scheme to avoid overfitting. Shahroudy {\it et al.} propose a part-aware LSTM network by separating the original LSTM cell into five sub-cells corresponding to five major groups of the human body \cite{Shahroudy_2016_CVPR}. Liu {\it et al.} propose a spatio-temporal LSTM structure to explore the contextual dependency of joints in both spatial and temporal domains \cite{liu2016spatio}. Li {\it et al.} propose an RNN tree network with a hierarchical structure to classify easy action classes at the lower layers and hard action classes at the higher layers \cite{li2017adaptive}. To address the large view variation of the captured data, Zhang {\it et al.} propose a view adaptive subnetwork which automatically determines the best observation viewpoints within an end-to-end network for recognition \cite{zhang2017view, zhang2019view}. N{\'u}{\~n}ez {\it et al.} \cite{nunez2018convolutional} combine CNN and LSTM together for action recognition and gesture recognition with a two stage training strategy. To explore the influence of contextual joints of one joint, Weng {\it et al.} \cite{weng2018deformable} propose a Deformable Pose Traversal Convolution scheme to traverse all the joints of a skeleton for both action recognition and gesture recognition. 

For RGB-based action recognition, convolutional neural networks are usually used to exploit the spatial dependencies \cite{simonyan2014two,wang2016temporal,yue2015beyond,donahue2015long}. Considering the temporal dependencies is also important for video, some approaches simply averaged/multiplied the scores or features of the frames for fusion \cite{simonyan2014two,wang2016temporal,diba2017deeptemporal}. Some other approaches leverage RNNs to model temporal correlations, with frame-wise CNN features as input at every time slot \cite{yue2015beyond,donahue2015long}. 

\subsection{Sequential MNIST Classification}
\label{sec:mnist}
MNIST classification is a digital recognition task, which has been widely used to evaluate the effectiveness of different algorithms \cite{lecun1998gradient}. Recently, Le {\it et al.} use the MNIST to evaluate the capability of modeling the long term dependencies of RNN structures by flattening the 784 pixels sequentially \cite{le2015simple}. It has become a widely used standard protocol to evaluate RNNs \cite{arjovsky2016unitary,cooijmans2016recurrent,campos2018skip,li2018independently}. 

Here we present some works which evaluate their improved RNNs on the sequential MNIST dataset. Le {\it et al.} propose a simple IRNN composed of rectified linear units which is initialized with the identity matrix to avoid gradient vanishing and exploding problems \cite{le2015simple}. Cooijmans {\it et al.} apply batch normalization to hidden-to-hidden transition to reduce the covariate shift among time steps \cite{cooijmans2016recurrent}. Li {\it et al.} treat all neurons in the same layer independently and build relationships between neurons at the next layer to address the gradient vanishing and exploding problems. \cite{li2018independently}.

Our proposed \EleAttG is a fundamental unit that aims to enhance the capability of an RNN block. We will demonstrate its effectiveness on different tasks, including 3D skeleton-based action recognition, RGB-based action recognition, gesture recognition, and sequential MNIST classification.
		
\section{Overview of Standard RNN, LSTM, and GRU}
\label{sec:RNN} 
Recurrent neural networks are capable of modeling temporal dynamics and spatial correlations of a time sequence. The key is the ``memory" mechanism which stores and updates historical information accumulated from previous time steps as time goes. To better understand the proposed EleAttG and its generalization capability, we briefly review the popular RNN structures, {\it i.e.,} standard RNN, and another two gate-based invariants, including Long Short Term Memory (LSTM) \cite{hochreiter1997long}, and Gated Recurrent Unit (GRU) \cite{cho14}:

\textbf{Standard RNN}. For a standard RNN layer, the output response \textbf{{h}$_t$} at time step $t$ is calculated from the input \textbf{{x}$_t$} of this layer and the output \textbf{{h}$_{t-1}$} of the last time slot as follows:
\begin{equation}
\label{equ:rnn}
{{\bf{h}}_{t}} = \tanh \left( {{\bf{W}}_{xh}}{{\bf{x}}_{t}} + {{\bf{W}}_{hh}}{{\bf{h}}_{t-1} + {\bf{b}}_h} \right),
\end{equation}
where ${\bf{W}}_{\alpha\beta}$ denotes the weight matrix related with $\alpha$ and $\beta$, where $\alpha$ $\in$ $\left\{ x, h \right\}$ and $\beta$ $\in$ $\left\{h \right\}$. $\bf{b}_{\gamma}$ is the bias vector, where $\gamma$ $\in$ $\left\{h \right\}$. 

\textbf{LSTM}. The standard RNN suffers from the gradient vanishing problem because of insufficient, decaying error back flow \cite{hochreiter1997long}. LSTM alleviates this problem by enforcing constant error flow through ``constant error carrousels" within the cell unit \textbf{{c}$_t$}. The input gate \textbf{{i}$_t$}, forget gate \textbf{{f}$_t$}, and output gate \textbf{{o}$_t$} open and close access to the constant error flow. For an LSTM layer, the recursive computations of activations of the units are as follows:
\begin{eqnarray}
\label{equ:lstm}
\begin{aligned}
\!&{\bf{i}}_t = \sigma \left( {{\bf{W}}_{xi}}{{\bf{x}}_{t}} + {{\bf{W}}_{hi}}{\bf{h}}_{t-1} + {\bf{b}}_i \right), \\ 
\!&{\bf{f}}_t = \sigma \left( {{\bf{W}}_{xf}}{{\bf{x}}_{t}} + {{\bf{W}}_{hf}}{\bf{h}}_{t-1} + {\bf{b}}_f \right), \\
\!&{\bf{c}}_t = {\bf{f}}_t\!\odot {\bf{c}}_{t-1}\!+{\bf{i}}_t \odot \tanh\! \left( {{\bf{W}}_{xc}}{{\bf{x}}_{t}}\! +\! {{\bf{W}}_{hc}}{\bf{h}}_{t-1} \!+\! {\bf{b}}_c \right), \\ 
\!&{\bf{o}}_t = \sigma \left( {{\bf{W}}_{xo}}{{\bf{x}}_{t}} + {{\bf{W}}_{ho}}{\bf{h}}_{t-1} + {\bf{b}}_o \right), \\
\!&{\bf{h}}_t = {\bf{o}}_t \odot \tanh \left( {\bf{c}}_t \right),
\end{aligned}
\end{eqnarray}
where $\odot$ denotes the element-wise product. Note that \textbf{{i}$_t$} is a vector representing the responses of a set of input gates of all LSTM neurons in this layer. \textbf{f$_t$}, \textbf{c$_t$}, \textbf{o$_t$}, and \textbf{h$_t$} are the output vectors of the forget gates, cells, output gates, hidden outputs of LSTM, respectively.

\textbf{GRU}. GRU is an architecture that is similar to but much simpler than LSTM. A GRU has two gates: reset gate \textbf{r$_t$} and update gate \textbf{z$_t$}. When the response of the reset gate is close to 0, the hidden state \textbf{h$_t'$} is forced to ignore the previous hidden state and reset with the current input only. The update gate controls how much information from the previous hidden state will be carried over to the current hidden state \textbf{h$_t$}. The hidden state acts in a similar role to the memory cell in LSTM. For a GRU layer, the recursive computations of activations of the units are as follows:
\begin{equation}
\label{equ:gru}
\begin{aligned}
\!&\mathbf{r}_t = \sigma \left( {\mathbf{W}}_{xr} \mathbf{x}_t + {\mathbf{W}}_{hr} \mathbf{h}_{t-1} + \mathbf{b}_r \right),\\
\!&\mathbf{z}_t = \sigma \left( {\mathbf{W}}_{xz} \mathbf{x}_t + {\mathbf{W}}_{hz} \mathbf{h}_{t-1} + \mathbf{b}_z \right),\\
\!&\mathbf{h}_t' = \tanh \left( {\mathbf{W}}_{xh} \mathbf{x}_t + {\mathbf{W}}_{hh} (\mathbf{r}_t \odot \mathbf{h}_{t-1}) + \mathbf{b}_h \right),\\
\!&\mathbf{h}_t =  {\mathbf{z_t}} \odot {\mathbf{h}_{t-1}}  + ({\mathbf{1 - z_t}}) \odot {\mathbf{h}_t'}.\\
\end{aligned}
\end{equation}
\textbf{r$_t$} and \textbf{z$_t$} are the output vectors of the reset gates, update gates of GRU. \textbf{h$_t$} denotes the output vector of the hidden state.

Note that in the weight matrices, {\it{e.g.},} \textbf{W$_{xr}$}, \textbf{W$_{xz}$} and\textbf{ W$_{xh}$} in Eq. (3),  the weights of the different dimensions of an input vector\textbf{ x$_t$} differ. However, for different samples, each weight matrix is shared.  The limitation of \textbf{W$_{xr}$}, W$_{xz}$ and \textbf{W$_{xh}$} is the lack of content adaptiveness. In contrast, our proposed EleAttG is able to control the contribution of different elements of the input depending on the input content.

\section{\Outer-Attention Gate for an RNN Block}
\label{sec:EleAtt-RNN}

For an RNN block, we propose an \Outer-Attention Gate (\EleAttGn) to empower the RNN neurons to have attentiveness capabilities. The response of an \EleAttG is a vector \textbf{{a}$_t$} with the same dimension as the input \textbf{{x}$_t$} of the RNNs, which is calculated as follows:
\begin{eqnarray}
\label{equ:agru}
\begin{aligned}
\!&\bf{a}_t =\phi \left({\bf{W}_{xa}}{\bf{x}_t} + {\bf{W}_{ha} {\bf{h}_{t-1}}} + {\bf{b}_a}  \right),
\end{aligned}
\end{eqnarray}
where $\phi$ denotes the activation function of Sigmoid, {\it i.e.,} $\phi(s) = 1/(1+e^{-s}$). The current input $\bf{x}_t$ and the previous hidden states \textbf{{h}$_{t-1}$} are used to determine the levels of importance of each element of the input \textbf{{x}$_t$}.

The attention response modulates the input to the updated input $\widetilde{\bf{x}_t}$ which is represented as: 
\begin{eqnarray}
\label{equ:updatedx}
\begin{aligned}
\!&\widetilde{\bf{x}_t}= \bf{a}_t \odot {\bf{x}_t}.
\end{aligned}
\end{eqnarray}
The recursive computations of activations of the other units in the RNN block are then based on the updated input $\widetilde{\bf{x}_t}$, instead of the original input \textbf{{x}$_t$}, as illustrated in Fig. \ref{fig:IAttRNN}.

\textbf{EleAtt-GRU}. For a GRU block with EleAttG (denoted as EleAtt-GRU), together with (\ref{equ:updatedx}), the computations for an EleAtt-GRU block are as follows.
\begin{equation}
\label{equ:EleAtt-gru}
\begin{aligned}
\!&\mathbf{r}_t = \sigma \left( {\mathbf{W}}_{xr} \widetilde{\bf{x}_t} + {\mathbf{W}}_{hr} \mathbf{h}_{t-1} + \mathbf{b}_r \right),\\
\!&\mathbf{z}_t = \sigma \left( {\mathbf{W}}_{xz} \widetilde{\bf{x}_t} + {\mathbf{W}}_{hz} \mathbf{h}_{t-1} + \mathbf{b}_z \right),\\
\!&\mathbf{h}_t' = \tanh \left( {\mathbf{W}}_{xh} \widetilde{\bf{x}_t} + {\mathbf{W}}_{hh} (\mathbf{r}_t \odot \mathbf{h}_{t-1}) + \mathbf{b}_h \right),\\
\!&\mathbf{h}_t =  {\mathbf{z_t}} \odot {\mathbf{h}_{t-1}}  + ({\mathbf{1 - z_t}}) \odot {\mathbf{h}_t'}.\\
\end{aligned}
\end{equation}

Similarly, we can get the recursive computations for EleAtt-sRNN and EleAtt-LSTM by setting $\bf{x_t}$ to be $\widetilde{\bf{x}_t}$ in (\ref{equ:rnn}) and (\ref{equ:lstm}).

Note that in our design, in an RNN block/layer, all neurons share the same \EleAttG (see (\ref{equ:updatedx}) and (\ref{equ:EleAtt-gru}) for the GRU block). Theoretically, each RNN neuron (instead of block) can have its own attention gate but the cost of computation complexity and number of parameters will largely increase, especially when the dimension of the input is large. We focus on the shared design in this work.

\section{Experiments}
We perform comprehensive studies to evaluate the effectiveness of our proposed \EARNN~with \EleAttG by applying it to action recognition from 3D skeleton data, and RGB video, gesture recognition, and sequential MNIST classification, respectively. 

To demonstrate the generalization capability of \EleAttGn, we add \EleAttG to the standard RNN, LSTM, and GRU structures, respectively. We also evaluate the effectiveness of \EleAttG on different types of input signals including skeleton data, CNN features, and raw image pixels.

For 3D skeleton-based action recognition we use three challenging datasets, {\it{i.e.},} the NTU RGB+D dataset (NTU) \cite{Shahroudy_2016_CVPR}, the Northwestern-UCLA dataset (N-UCLA) \cite{wang2014cross}, and the SYSU Human-Object Interaction dataset (SYSU)\cite{hu2015jointly}. The NTU dataset is currently the largest dataset with diverse subjects, various viewpoints and small inter-class differences. Therefore, our in-depth analyses are performed on the NTU dataset. For RGB-based action recognition, we take the CNN features extracted from existing, pre-trained models without finetuning as the input to the RNN-based recognition networks and evaluate the effectiveness of \EleAttG on the RGB videos of the NTU and the JHMDB datasets \cite{jhuang2013towards}. For gesture recognition, we use the Dynamic Hand Gesture 14/28 (DHG) dataset \cite{de2016skeleton}. For sequential MNIST classification, we use the MNIST handwritten digits benchmark dataset \cite{lecun1998gradient}.  We conduct most of our experiments based on GRU, as it has a simpler structure than LSTM and better performance than the standard RNN.

\begin{figure*}[!]
	\centering
	\begin{subfigure}[t]{0.32\linewidth}
		\centering\includegraphics[width=\textwidth]{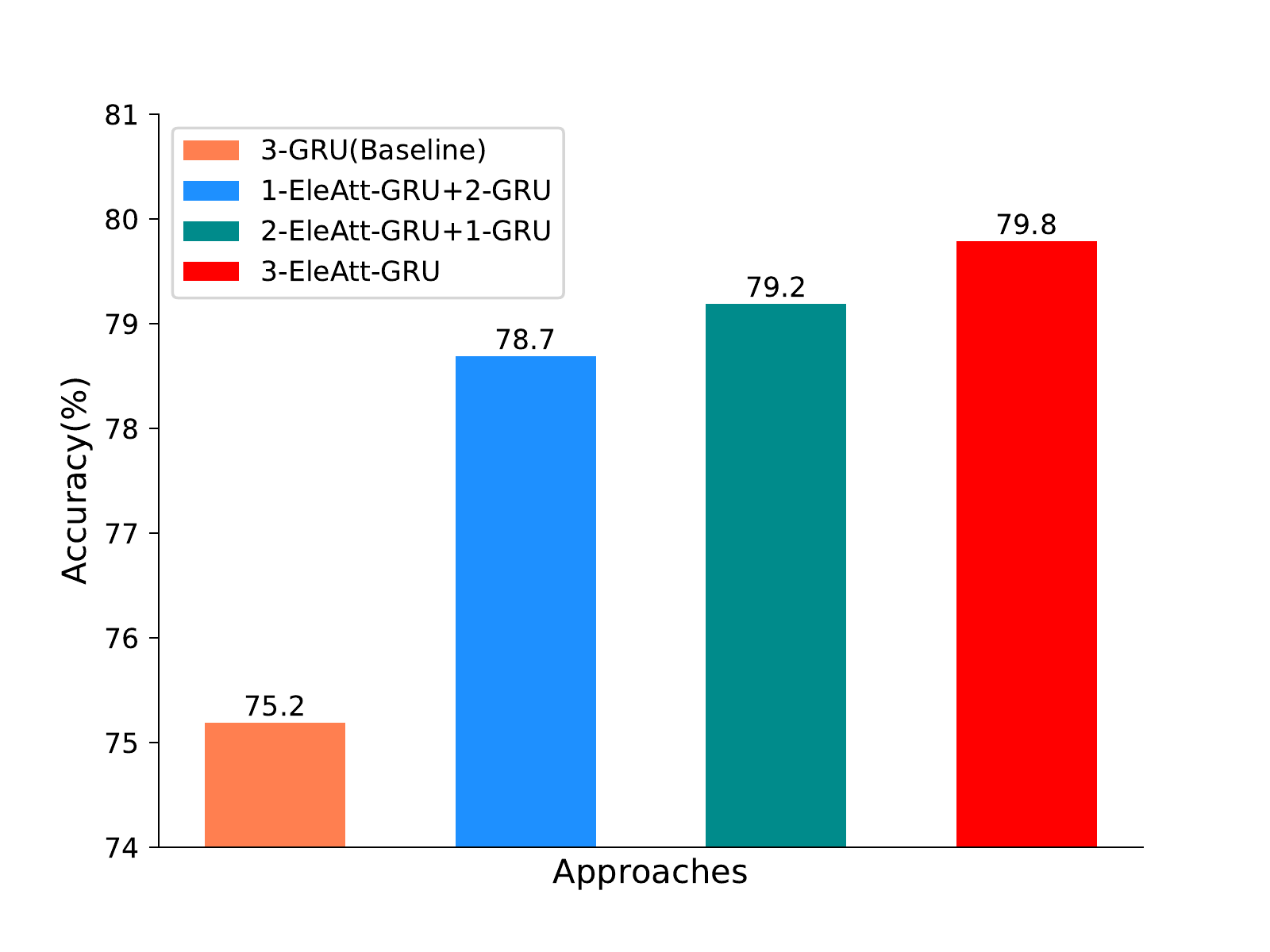}
		\caption{NTU-CS}
		\label{subfig:CS}
	\end{subfigure}	
	\hfil
	\begin{subfigure}[t]{0.32\linewidth}
		\centering\includegraphics[width=\textwidth]{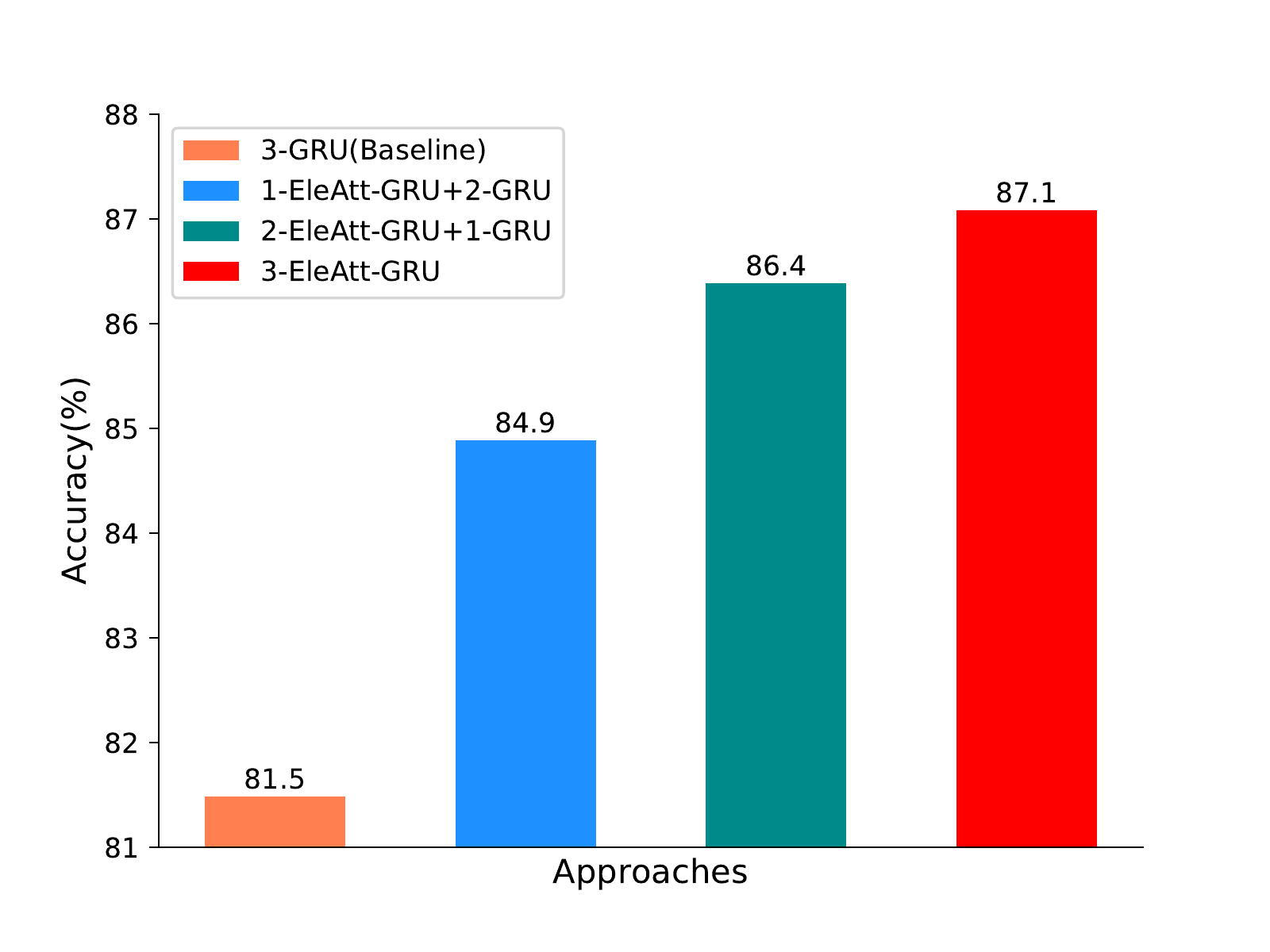}
		\caption{NTU-CV}			
		\label{subfig:CV}
	\end{subfigure}
	\hfil
	\begin{subfigure}[t]{0.32\linewidth}
		\centering\includegraphics[width=\textwidth]{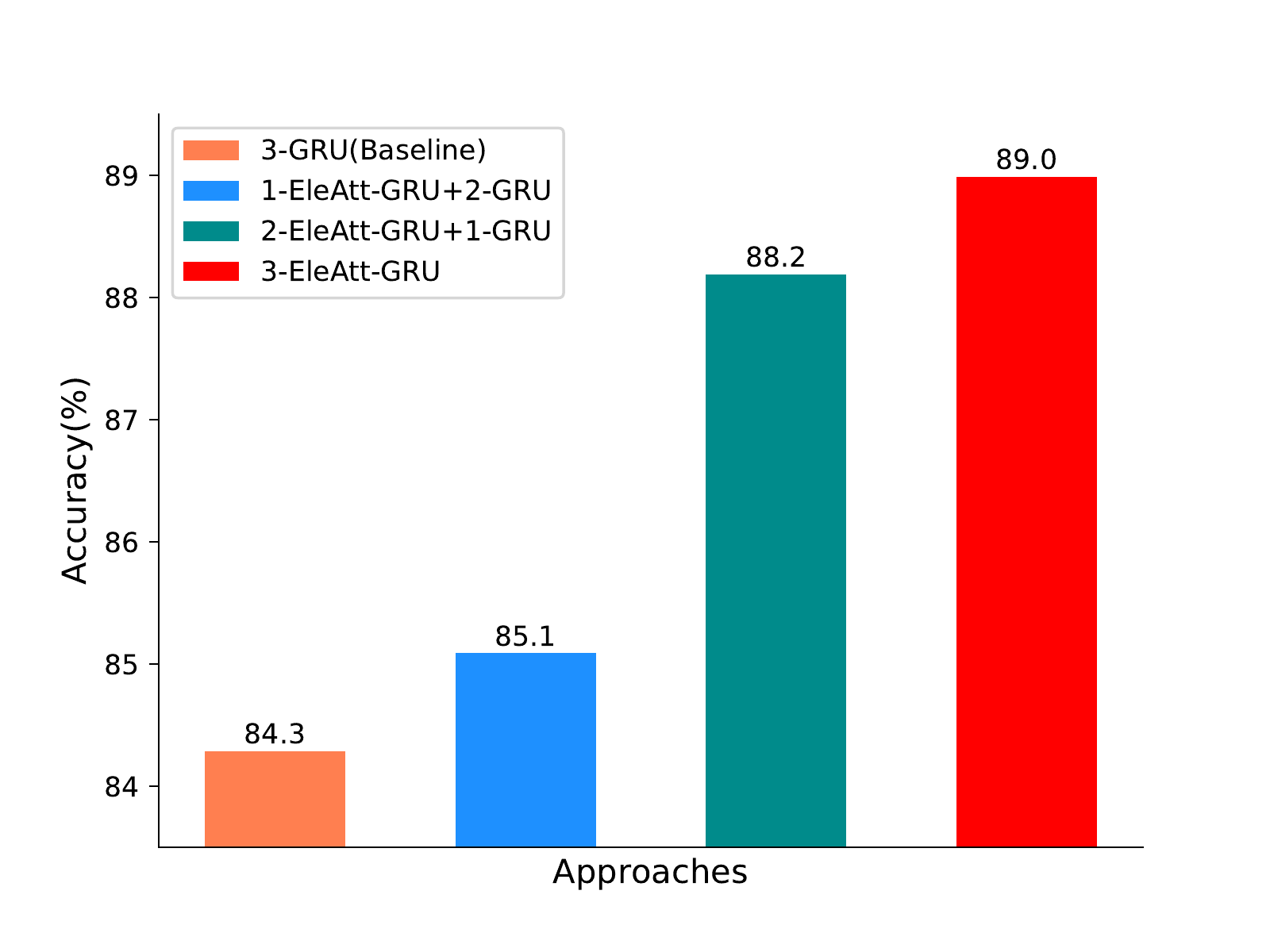}
		\caption{N-UCLA}			
		\label{subfig:N-UCLA}
	\end{subfigure}
	\caption{Effectiveness of our proposed EleAttG on the three-layer GRU network for 3D skeleton-based human action recognition for the CS and CV settings of the NTU dataset, and the N-UCLA dataset. ``$m$-EleAtt-GRU+$n$-GRU" denotes that the first $m$ layers are EleAtt-GRU layers and the remaining $n$ layers are the original GRU layers.}
	\label{fig:layer}
\end{figure*}

\subsection{Datasets}
\label{datasets}
\textbf{NTU RGB+D Dataset (NTU)~\cite{Shahroudy_2016_CVPR}.} NTU is currently the largest RGB+D+Skeleton dataset for action recognition, which includes 56880 videos with more than 4 million frames in total. There are 60 kinds of action classes performed by 40 subjects. Each subject has 25 body joints and each joint has 3D coordinates. Three cameras placed in different positions are used to capture the data at the same time and there are over 80 views. We follow the standard protocols proposed in \cite{Shahroudy_2016_CVPR} including the Cross Subject (CS) and Cross View (CV) settings. For the CS setting, 40 subjects are equally split into training and testing groups. For the CV setting, the samples of cameras 2 and 3 are adopted for training while those of camera 1 are for testing. 

\textbf{Northwestern-UCLA dataset (N-UCLA)~\cite{wang2014cross}.} N-UCLA is a small RGB+D+Skeleton dataset including 1494 sequences performed by 10 subjects. It records 10 different actions in total. 20 joints with 3D coordinates are provided for each human body in this dataset. Following \cite{wang2014cross}, we use samples from the first two cameras as training data, and the samples from the third camera as testing data. 

\textbf{SYSU Human-Object Interaction dataset (SYSU)~\cite{hu2015jointly}.} SYSU is a small RGB+D+Skeleton dataset, including 480 sequences performed by 40 different subjects. It contains a total of 12 actions. A subject has 20 joints with 3D coordinates. We follow the standard protocols proposed in \cite{hu2015jointly} for evaluation. They include two settings. For the Cross Subject (CS) setting, half of the subjects are used for training and the others for testing. For the Same Subject (SS) setting, half of the sequences of each subject are used for training and others for testing. The average performance of 30-fold cross validation is reported.

\textbf{JHMDB dataset (JHMDB)~\cite{jhuang2013towards}.} JHMDB is an RGB-based dataset which has 928 RGB videos with each video containing about 15-40 frames. It contains 21 actions performed by different actors. This dataset is challenging where the videos are collected on the Internet which also includes outdoor activities.

\textbf{Dynamic Hand Gesture 14/28 dataset (DHG)~\cite{de2016skeleton}.} DHG is an Intel Real Sense captured dataset for gesture recognition, including 2800 sequences performed by 20 subjects. Each gesture is represented by 22 joints with 3D coordinates. It contains 14 classes of gestures and each gesture is collected in two ways: using the whole hand or just one finger. We follow the standard leave-one-subject-out cross-validation strategy under two evaluation protocols: 14 classes or 28 classes which treats the same gesture performed by one finger or whole hand separately \cite{de2016skeleton}.

\textbf{MNIST dataset (MNIST) \cite{lecun1998gradient}.} MNIST is a handwritten digit dataset which is widely used to evaluate different methods, such as CNN and RNN. In order to utilize MNIST to evaluate RNN structure, we flat the images with size of 28 $\times$ 28 to 784-dimension vectors following \cite{le2015simple,arjovsky2016unitary,cooijmans2016recurrent,campos2018skip,li2018independently}. RNN treats each dimension as the input of a time slot. We follow the standard data split protocol for training and testing, containing 60000 training samples and 10000 testing samples. In addition, we randomly select 5000 samples from training sets for validation.

\subsection{Implementation Details}
\label{subsec:implement}

We perform our experiments on the deep learning platform of Keras \cite{chollet2015keras} with Theano \cite{al2016theano} as the backend. Dropout \cite{srivastava2014dropout} with the probability of 0.5 is used to alleviate overfitting. Gradient clipping similar to \cite{sutskever2014sequence} is used by constraining the maximum amplitude of the gradient to 1. Adam \cite{kingma2014adam} is used to train the networks from end-to-end. The initial learning rate is set to 0.005 for 3D skeleton-based action recognition, gesture recognition, and sequential MNIST classification and 0.001 for RGB-based action recognition. During training, the learning rate will be reduced by a factor of 10 when the training accuracy does not increase for some epochs. We use cross-entropy as the loss function to train all the networks. 

For 3D skeleton-based action recognition and gesture recognition, similar to the classification network design in \cite{zhang2017view}, we build our recognition networks by stacking three RNN layers with {\EleAttGn}s and one fully connected (FC) layer for classification. We use 100 RNN neurons in each layer. Considering the large differences in the size of the datasets, we set the batch size for the NTU, N-UCLA, SYSU and DHG datasets to 256, 32, 32, and 32, respectively. We use the sequence-level pre-processing method in \cite{zhang2017view} by setting the body center in the first frame as the coordinate origin to make the system invariant to the initial position of human body. To improve the robustness to view variations at the sequence level, we perform data augmentation by randomly rotating the skeleton around the X, Y and Z axes by various degrees ranging from -17 to 17 during training for 3D skeleton-based action recognition. For the N-UCLA and SYSU datasets, we use the RNN models pre-trained on a sub NTU dataset, where each subject has 20 joints and only the actions performed by one subject are used, to initialize the baseline schemes and the proposed schemes.   

For RGB-based action recognition, we feed an RNN network with the CNN features to further explore temporal dynamics. Because our aim is to evaluate whether the proposed \EleAttG can generally improve recognition accuracy in different types of input signals, we extract CNN features using some available pre-trained models without finetuning for the specific dataset or task. For the JHMDB dataset, we use the TSN model from \cite{wang2016temporal,TSNModel} which was trained on the HMDB dataset \cite{kuehne2011hmdb} to extract a 1024 dimensional feature for each frame. For the NTU dataset which has more videos, we take the ResNet50 model \cite{he2016deep,ResNet50Model} which has been pre-trained on ImageNet as our feature extractor (2048 dimensional feature for each frame). The implementation details of the RNN networks are similar to those discussed above. For the NTU dataset, we stack three EleAtt-GRU layers, with each layer consisting of 512 GRU neurons. For the JHMDB dataset, we use only one GRU layer (512 GRU neurons) with \EleAttG to avoid overfitting, considering that the number of video samples is much smaller than that of the NTU dataset. We set the batch size for the NTU, and JHMDB datasets to 256 and 32, respectively.

For sequential MNIST classification, we build the same classification network as the skeleton-based recognition network mentioned above, i.e. stacking three RNN layers with {\EleAttGn}s and one fully connected layer (FC). We set the batch size to 256.

\subsection{Effectiveness of \Outer-Attention-Gates}

In this section, we first demonstrate the effectiveness of the proposed \EleAttG. Then we demonstrate good generalization performance on different RNN structures and on different types of signals. Finally, comparisons with the state-of-the-art approaches are performed followed by the visualization of the learned attention.

\textbf{Effectiveness on GRU network.} We evaluate the performance on 3D skeleton-based action recognition on the NTU and N-UCLA datasets. Fig. \ref{fig:layer} shows the effectiveness of {\EleAttGn} on the baseline network which consists of 3 GRU layers. We find that our final scheme with three EleAtt-GRU layers (``3-EleAtt-GRU") outperforms the baseline scheme (``3-GRU(Baseline)")  by {\bf{4.6\%}}, {\bf{5.6\%}}, and {\bf{4.7\%}, for the NTU-CS, NTU-CV, and N-UCLA}, respectively. However, there is something different between the NTU dataset and N-UCLA dataset. \textbf{1)} For the N-UCLA dataset, the improvement of ``1-EleAtt-GRU+2-GRU" in comparison with ``3-GRU(Baseline)"  is not as obvious as that of the NTU dataset. \textbf{2)} The improvement when increasing one GRU layer with EleAttG to two GRU layers with EleAttGs is significant in the N-UCLA dataset, {\it i.e.}, 3.1\% in accuracy. The potential reason is caused by the difference of action class distributions of the two datasets. We find that the diversity of action classes of the N-UCLA dataset (10 classes) is much smaller than that of the NTU dataset (60 classes). The important joint tends to be related with hands for most actions which can be learned by the dataset level weights. Therefore, the gain by using EleAttG of the first layer is smaller for the N-UCLA dataset. For the input features of the second layer, our EleAttG can help to better adaptively capture the discriminative feature dimensions and improve performance. Similar phenomena are also observed on the SYSU dataset which has less action class diversity.

The overall trend is same for both datasets. The performance grows when more GRU layers with EleAttGs are used. This indicates that, besides the attention on skeleton joints, the suitable attention on features can also significantly improve performance.

\textbf{Generalization to other input signals.} The proposed RNN block with {\EleAttGn} is generic and can be applied to different types of source data. To demonstrate this, we use (1) CNN features extracted from RGB frames as the input of the RNNs for RGB based action recognition, and (2) raw image pixels as the inputs of the RNNs for sequential MNIST classification. 

For RGB-based action recognition, Table \ref{tab:rgb} shows the performance comparisons on the NTU and JHMDB datasets. The implementation details have been described in Section \ref{subsec:implement}. The ``EleAtt-GRU" outperforms the ``Baseline-GRU" by about 2-4\% on the NTU dataset, and 2\% on the JHMDB dataset. Note that the performance is not optimized since we have not used the fine-tuned CNN model on this dataset for this task.

For sequential MNIST classification, Table \ref{tab:MNIST} shows the performance comparison on the sequential MNIST dataset, where the inputs are raw pixels. The implementation details have been described in Section \ref{subsec:implement}. The ``EleAtt-GRU" is superior to the ``Baseline-GRU" by 0.4\%. Note that it is difficulty to further improve the performance significantly since the final accuracy is already very high.

\setlength{\tabcolsep}{7pt}
\begin{table}[t]
	\centering
	\caption{Effectiveness of {\EleAttGn}s in the GRU network for RGB-based action recognition on the NTU and JHMDB datasets. Here, CNN features for each RGB frame is taken as the input to the GRU network. The performance is evaluated in terms of recognition accuracy(\%).}
	\label{tab:rgb}
	\begin{tabular}{ccccccc}
		\toprule
		\multirow{2}{*}{Dataset} & \multicolumn{2}{c}{NTU} & \multicolumn{4}{c}{JHMDB}        \\
		\cmidrule(lr){2-3}
		\cmidrule(lr){4-7}
		& CS         & CV         & Split1 & Split2 & Split3 & Average  \\
		\midrule
		Baseline-GRU                & 61.3      & 66.8      & 60.6  & 59.2  & 62.9  & 60.9 \\
		EleAtt-GRU                    & 63.3      & 70.6      & 64.5  & 59.2  & 65.0  & 62.9 \\
		\bottomrule
	\end{tabular}
\end{table}

\setlength{\tabcolsep}{7pt}
\begin{table}[t]
	\centering
	\caption{Effectiveness of {\EleAttGn} on three types of RNN structures. We evaluate the recognition performance of RNN networks with {\EleAttGn}s for the CS and CV settings of the NTU dataset. ``EleAtt-$X$" denotes the scheme with {\EleAttGn}s based on the RNN structure of $X$.}
	\label{tab:Extend}
	\begin{tabular}{cccc}
		\toprule
		RNN structure                    & Scheme     & CS & CV \\
		\midrule
		\multirow{2}{*}{Standard RNN} & Baseline(1-sRNN)   & 51.6    & 57.6 \\
		& EleAtt-sRNN & \textbf{61.6}     & \textbf{67.2}  \\
		\midrule
		\multirow{2}{*}{LSTM}      & Baseline(3-LSTM)   & 77.2     & 83.0  \\
		& EleAtt-LSTM     & \textbf{78.4}     & \textbf{85.0}  \\
		\midrule
		\multirow{2}{*}{GRU}       & Baseline(3-GRU)   & 75.2     & 81.5  \\
		& EleAtt-GRU       & \textbf{79.8}     & \textbf{87.1} 							\\
		\bottomrule	
	\end{tabular}
\end{table}

\setlength{\tabcolsep}{7pt}
\begin{table}[t]
	\centering
	\caption{Performance comparisons on the NTU dataset in terms of accuracy (\%).}
	\label{tab:ntu}
	\begin{tabular}{ccc}
		\toprule
		{Method}                                           & CS & CV \\
		\midrule
		{Skeleton Quads \cite{evangelidis2014skeletal}}  & 38.6     & 41.4  \\
		{Lie Group \cite{vemulapalli2014human}}          & 50.1     & 52.8  \\
		{Dynamic Skeletons  \cite{hu2015jointly}}        & 60.2    & 65.2  \\
		{HBRNN-L  \cite{du2015hierarchical}}             & 59.1     & 64.0  \\
		{Part-aware LSTM   \cite{Shahroudy_2016_CVPR}} & 62.9     & 70.3  \\
		{ST-LSTM + Trust Gate \cite{liu2016spatio}}      & 69.2     & 77.7  \\
		{STA-LSTM \cite{song2018spatio}}                    & 73.4     & 81.2  \\
		{GCA-LSTM \cite{liu2017global}}                  & 74.4     & 82.8  \\
		{URNN-2L-T \cite{li2017adaptive}}                & 74.6     & 83.2  \\
		{Clips+CNN+MTLN \cite{ke2017new}}                & 79.6     & 84.8  \\
		{VA-LSTM \cite{zhang2017view}}                   & 79.4     & 87.2  \\
		\midrule
		Baseline-GRU  & 75.2 & 81.5 \\
		EleAtt-GRU & 79.8  & 87.1 \\
		EleAtt-GRU(aug.)     & \textbf{80.7}    & \textbf{88.4} \\
		\bottomrule
	\end{tabular}
\end{table}

\textbf{Generalization on various RNN structures.} The proposed \EleAttG is generic and can be applied to various types of RNN structures. We evaluate the effects of {\EleAttGn}s on three classical RNN structures, {\it i.e.,} the standard RNN (sRNN), LSTM, and GRU respectively and show the results in Table \ref{tab:Extend}. Compared with LSTM and GRU, the standard RNN neurons do not have the gating designs which control the contributions of the current input to the network. The \EleAttG can element-wisely control the contribution of the current input, which remedies the lack of gate designs to some extent. The gate designs in LSTM and GRU can only control the information flow input-wisely. In contrast, the proposed {\EleAttGn}s are capable of modulating the input element-wisely, which empowers the attentiveness capability to RNNs. We can see that the adding of {\EleAttGn}s  enhances performance significantly. Note that for sRNN, we build both the ``Baseline(1-sRNN)" and our scheme using only one sRNN layer rather than three as those for LSTM and GRU, in considering that the three-layer sRNN baseline converges to a poorer performance, {\it i.e.}, 33.6\% and 42.8\% for the CS and CV settings, which may be caused by the gradient vanishing of sRNN. 

\setlength{\tabcolsep}{7pt}
\begin{table}[!t]
	\centering
	\caption{Performance comparisons on the N-UCLA dataset in terms of accuracy (\%).}
	\begin{tabular}{ccc}
		\toprule
		Method & Accuracy \\
		\midrule
		HOJ3D \cite{xia2012view}          & 54.5      \\ 
		AE  \cite{wang2013learning}            & 76.0    \\ 
		HBRNN-L \cite{du2016representation}        & 78.5     \\ 
		DA-Net \cite{wang2018dividing}   &86.5 \\
		\midrule
		Baseline-GRU & 84.3 \\
		EleAtt-GRU & 89.0 \\
		EleAtt-GRU(aug.) & \bf{90.7}\\ 
		\bottomrule     
		\label{tab:N-UCLA}
	\end{tabular}
\end{table}

\setlength{\tabcolsep}{7pt}
\begin{table}[!t]
	\centering
	\caption{Performance comparisons on the SYSU dataset in terms of accuracy (\%).}
	\begin{tabular}{ccc}
		\toprule
		Method & SS  & CS \\
		\midrule
		LAFF \cite{hu2016real} & - & 54.2 \\
		DS \cite{hu2015jointly} & 75.5 & 76.9 \\	
		VA-LSTM \cite{zhang2017view} & 76.9 & 77.5 \\
		SR-TSL \cite{si2018skeleton} \  &80.7 &81.9 \\
		\midrule
		Baseline-GRU  &  82.1  & 82.1 \\
		EleAtt-GRU & 84.9 & 84.5 \\
		EleAtt-GRU(aug.) & \bf{85.7} & \bf{85.7} \\
		\bottomrule      
		\label{tab:SYSU}
	\end{tabular}
	
\end{table}

\textbf{Comparisons with state-of-the-arts on skeleton-based action recognition, gesture recognition, and sequential MNIST classification.} For 3D skeleton-based human action recognition, a great deal of approaches have been proposed for enhancing the recognition accuracy as discussed in Section \ref{subsec:action}. To achieve good performance, some approaches require complicated designs \cite{liu2016spatio,liu2017global} while some others are specially designed considering human body characteristics \cite{du2015hierarchical,Shahroudy_2016_CVPR,song2017end}. In contrast, our proposed {\EleAttGn}s for RNN blocks are not specially designed for human body signals and can be used for other tasks directly. 

\begin{figure*}[th]
	\centering
	\includegraphics[width=0.8\linewidth]{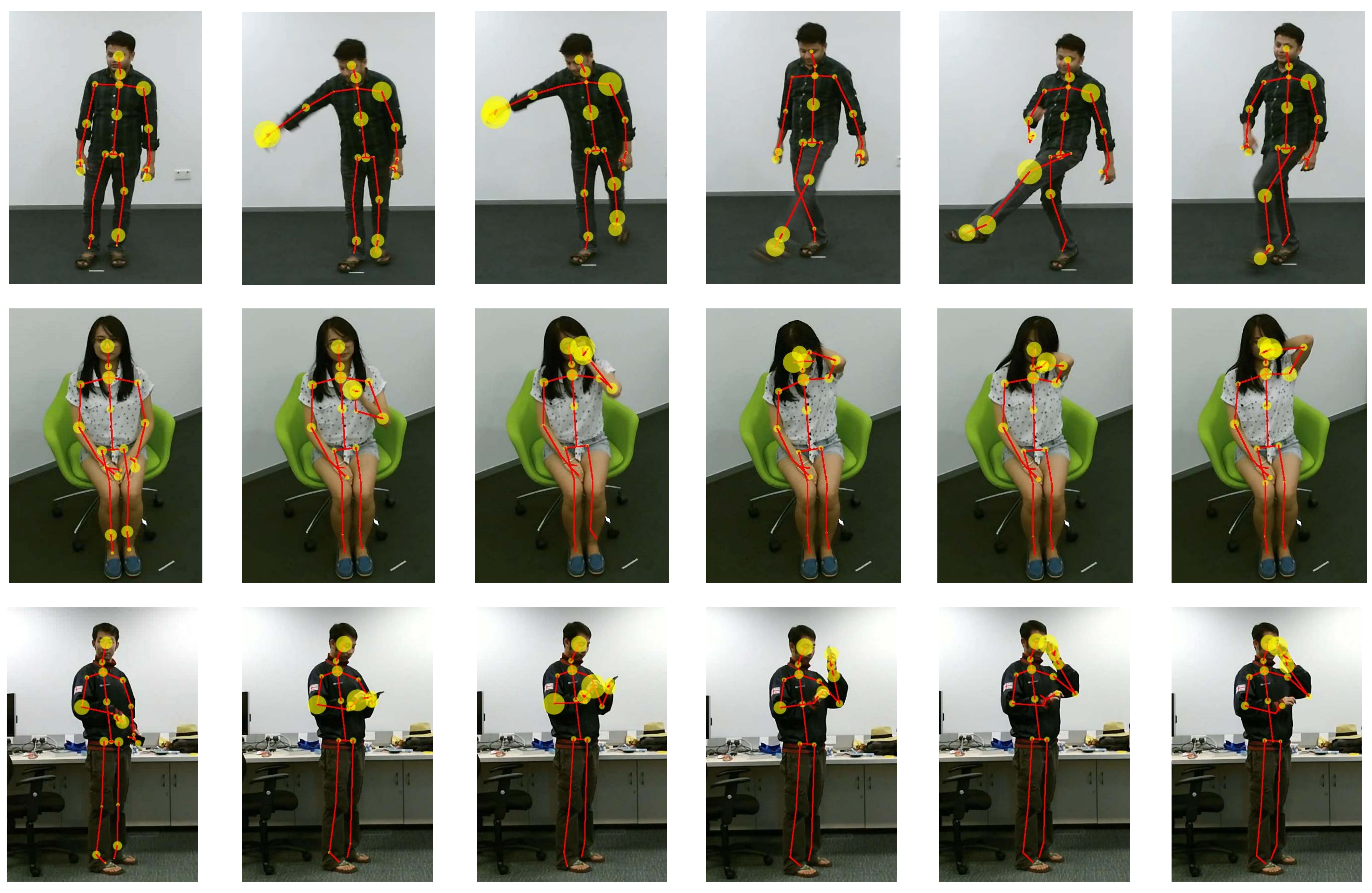}
	\caption{Visualization based on the attention responses of the first GRU layer for the actions of \emph{kicking}, \emph{touching neck} and \emph{making a phone call}. For each joint, the size of the yellow circle indicates the learned level of importance. Here, the levels of importance for the $X$, $Y$, $Z$ coordinates of a joint are summed for visualization.}	
	\label{fig:vis}
\end{figure*}

\begin{figure}[!]
	\centering
	\includegraphics[width=0.3\linewidth]{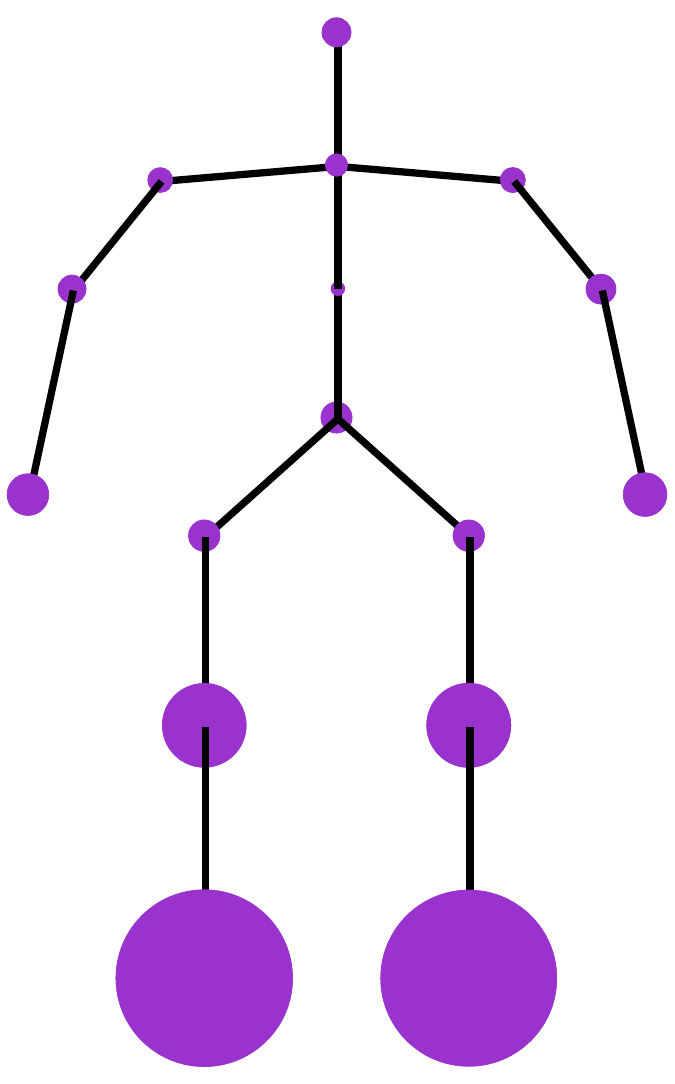}
	\caption{Illustration of the statistical energy of each joint. The energy of one joint is the summation of the statistical energies of its three elements, {\it i.e., $X$, $Y$, and $Z$} coordinates. The size of the circle on each joint is proportional to the energy of that joint. The larger of the circle size, the larger of the energy. We only show the main joints of human body for clarify.}	
	\label{fig:energy}
\end{figure}

\begin{figure*}[th]
	\centering
	\includegraphics[width=0.8\linewidth]{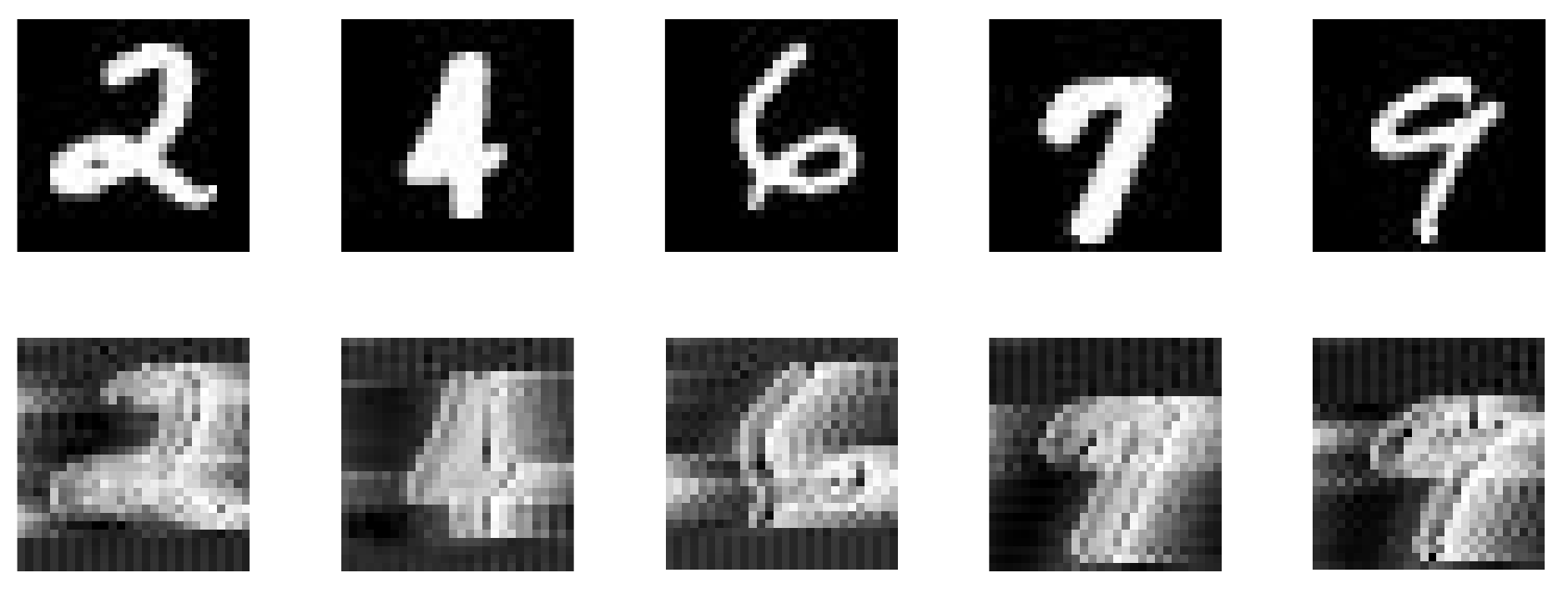}
	\caption{Visualization based on the attention responses of the first GRU layer for different handwritten digits. The first row denotes the original digit images and the second row denotes the corresponding attention responses of images pixels. Note that we use larger value to denote larger attention response value.}	
	\label{fig:mnist}
\end{figure*}

For action recognition, Tables \ref{tab:ntu}, \ref{tab:N-UCLA}, and \ref{tab:SYSU} show the performance comparisons with state-of-the-art approaches on the NTU, N-UCLA and SYSU datasets, respectively. ``Baseline-GRU" denotes our baseline scheme which is built by stacking three GRU layers while ``EleAtt-GRU" denotes our proposed scheme which replaces the GRU layers by the proposed GRU layers with {\EleAttGn}s. Implementation details can be found in Section \ref{subsec:implement}.  ``EleAtt-GRU(aug.)" denotes that data argumentation by rotating skeleton sequences is performed during training. We achieve the best performance in comparison with other state-of-the-art approaches on all three datasets. Our scheme ``EleAtt-GRU" achieves significant gains over the baseline scheme ``Baseline-GRU", of 4.6-5.6\%, 4.7\%, and 2.4-2.8\% on the NTU, N-UCLA, and SYSU datasets, respectively.

For gesture recognition, even without any special designs for gesture like \cite{nunez2018convolutional} and \cite{weng2018deformable}, we achieve the best results as shown in Table \ref{tab:DHG}. Our proposed ``EleAtt-GRU" outperforms the current best method by 5.2\% and 6.3\% for the ``C=14" and ``C=28" settings, respectively. 

For sequential MNIST classification, we achieve the best performance in comparison with the state-of-the-arts as shown in Table \ref{tab:MNIST}. Additionally, our scheme ``EleAtt-GRU" is superior to the baseline scheme ``Baseline-GRU" by 0.4\%.

\setlength{\tabcolsep}{7pt}
\begin{table}[!]
	\centering
	\caption{Performance comparisons on the gesture recognition dataset DHG in terms of accuracy (\%).}
	\begin{tabular}{ccc}
		\toprule
		Method & C=14  & C=28 \\
		\midrule
		Skeleton Quads \cite{evangelidis2014skeletal} & 84.5  & 79.4  \\
		SoCJ+HoHD+HoWR \cite{de2016skeleton} & 83.1  & 80.0  \\
		CNN+LSTM \cite{nunez2018convolutional} & 85.6  & 81.1  \\
		D-Pose Traversal Conv \cite{weng2018deformable} & 85.8  & 80.2  \\
		\midrule
		Baseline-GRU & 90.0  & 85.9  \\
		EleAtt-GRU & \textbf{91.0}  & \textbf{86.5}  \\
		\bottomrule
	\end{tabular}%
	\label{tab:DHG}%
\end{table}%
\setlength{\tabcolsep}{7pt}
\begin{table}[t]
	\centering
	\caption{Performance comparisons on the sequential MNIST dataset in terms of accuracy (\%).}
	\begin{tabular}{cc}
		\toprule
		Method & Accuracy \\
		\midrule
		TANH-RNN \cite{le2015simple} & 35.0  \\
		iRNN \cite{le2015simple}  & 97.0  \\
		uRNN \cite{arjovsky2016unitary}  & 95.1  \\
		sTANH-RNN \cite{zhang2016architectural}& 98.1  \\
		LSTM \cite{cooijmans2016recurrent}  & 98.9  \\
		BN-LSTM \cite{cooijmans2016recurrent}& 99.0  \\
		Skip GRU \cite{campos2018skip}& 97.6  \\
		Skip LSTM \cite{campos2018skip}& 97.3  \\
		IndRNN (6 layers) \cite{li2018independently} & 99.0  \\
		\midrule
		Bseline-GRU & 98.8  \\
		EleAtt-GRU & \textbf{99.2}  \\
		\bottomrule
	\end{tabular}%
	\label{tab:MNIST}%
\end{table}%

\textbf{Visualization of the responses of \EleAttGn.} To better understand the learned element-wise attention, we observe the responses of the \EleAttG in the first GRU layer for the skeleton-based action recognition and sequential MNIST classification.

\begin{figure*}[!]
	\centering
	\begin{subfigure}[t]{0.41\linewidth}
		\centering\includegraphics[width=\textwidth]{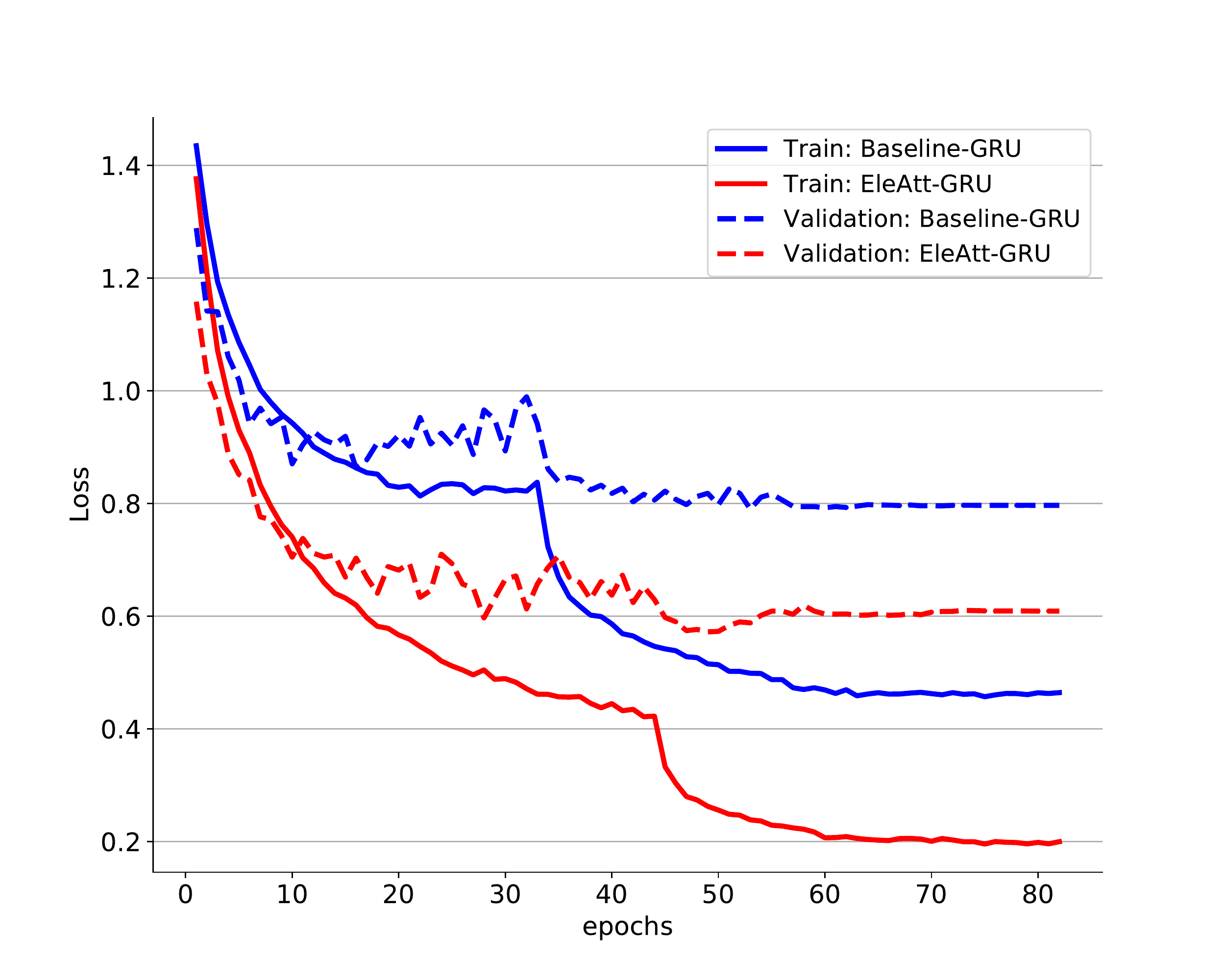}
		\caption{NTU-CS}
		\label{subfig:CS}
	\end{subfigure}	
	\hfil
	\begin{subfigure}[t]{0.41\linewidth}
    	\centering\includegraphics[width=\textwidth]{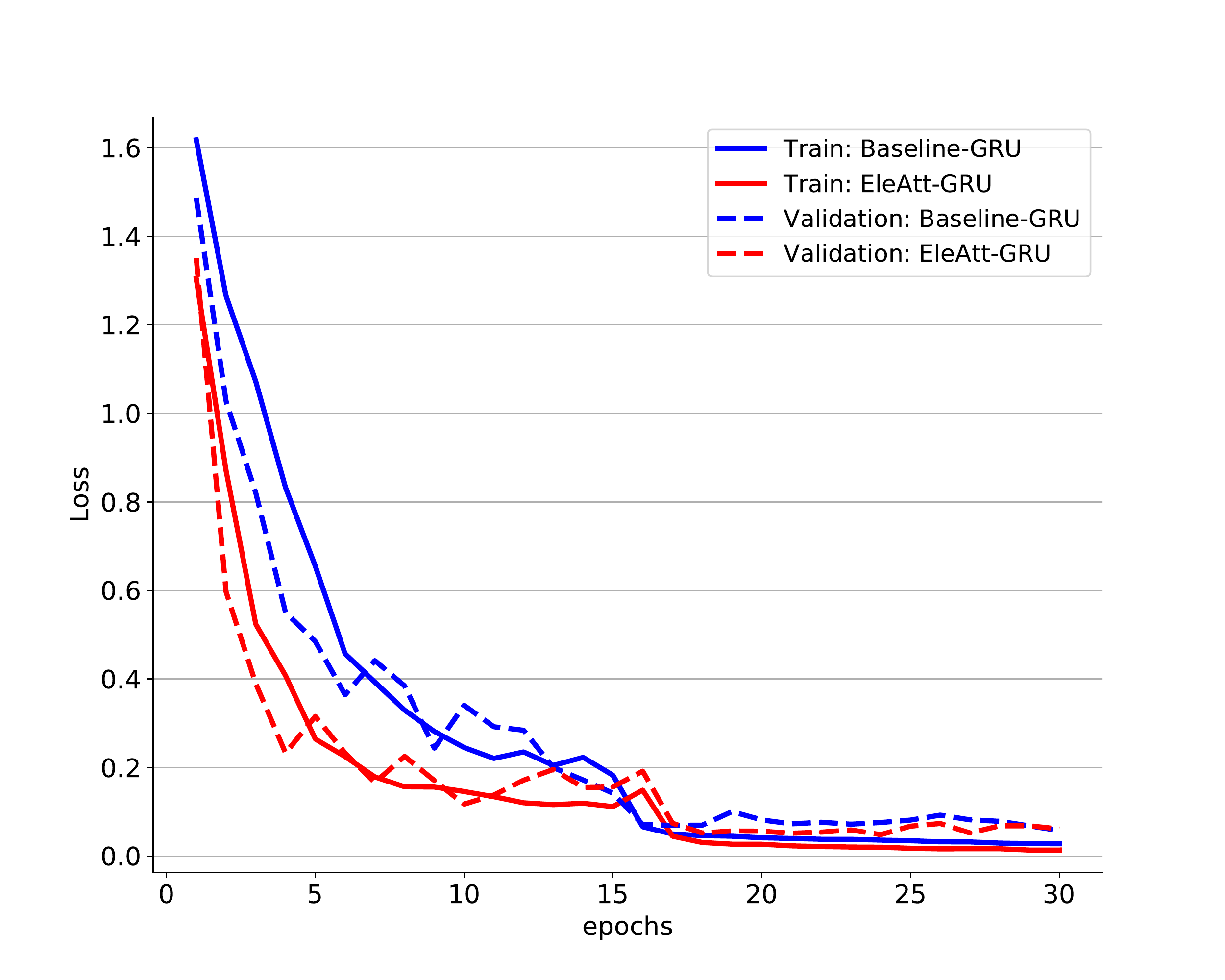}
    	\caption{MNIST}			
    	\label{subfig:CV}
    \end{subfigure}
	\caption{Loss curves during training on (a) the CS setting of the NTU dataset, and (b) the sequential MNIST dataset with respect to the proposed scheme ``EleAtt-GRU" and the baseline scheme ``Baseline-GRU".}
	\label{fig:curve}
\end{figure*}

For skeleton-based action recognition, in the first layer, the input (with dimension of $3\times J$) at a time step has $J$ joints and each joint is represented by the $X$, $Y$, and $Z$ coordinate values. Apparently, the physical meaning of the attention responses in the first layer is clear. However, in a higher layer, the \EleAttG modulates the input features on each element which is more difficult to interpret and the attention value is hard to visualize. Therefore, we perform visualization based on the attention responses of the first GRU layer in Fig.~\ref{fig:vis} for the actions consisting of \emph{kicking}, \emph{touching the neck} and \emph{making a phone call}. 

Actually, the absolute response values cannot represent the relative importance of different elements of input very well. The statistical energies of different elements of the original input are different. The statistical energy of one element denotes the mean of energy of all skeleton sequences, where the energy is defined as the square of the value of one element. Considering each joint has three elements (3D coordinates), we denote the sum of the statistical energies of its three elements as the statistical energy of one joint. We illustrate the statistical energy of each joint in Fig \ref{fig:energy}. For example, the foot joint, which is in general far away from the body center, has a higher energy than that of the body center joint. We can imagine that there is a static modulation $\overline{a_i}$ on the $i^{th}$ element of the input, which can be calculated by the energy before and after the modulation. For the $i^{th}$ element of an sample $j$ with attention value $a_{i,j}$, we use the relative response value $\widehat{a_{i,j}} = a_{i,j}/\overline{a_i}$ for visualization to better reflect importance among joints. Note that the sum of the relative responses for the $X$, $Y$, and $Z$ of a joint is utilized for visualization. For the actions of \emph{touching neck} and \emph{making a phone call}  which are highly concerned with the joints on the arms and heads, thus, the relative attention on those joints are larger. For \emph{kicking}, the relative attention on the legs is large. These are consistent with a human's perception. 
For sequential MNIST classification, similar to the skeleton-based action recognition, we only perform visualization based on the attention responses of the first GRU layer and show some results in Fig. \ref{fig:mnist} for the same reason. Different from skeleton-based action recognition, we directly use the attention response value $a_{i,j}$ for visualization. Note that the input to the first layer at each time slot corresponds to the pixel values of a position and the pixels are scanned pixel by pixel. The response value of the {\EleAttGn} reflects the importance level of the corresponding image pixel directly very well. From Fig. \ref{fig:mnist}, we find that the pixels with respect to the handwritten digit are more important than other pixels. 

\subsection{Discussions}
\textbf{Training curves on the NTU and MNIST datasets.} Fig. \ref{fig:curve} (a) shows the training loss curves for the training and validation sets on the NTU dataset with CS setting during the training precedure for the proposed ``EleAtt-GRU" and the baseline ``Baseline-GRU", respectively. By adding the {\EleAttGn}s, ``EleAtt-GRU" takes fewer epochs than ``Baseline-GRU" to achieve the same training and validation losses, especially at the beginning of training. In addition, ``EleAtt-GRU" is consistently better than the ``Baseline-GRU". Similar phenomenon can be found on the sequential MNIST dataset in Fig. \ref{fig:curve} (b). The modulation of input can control the information flow of each input element adaptively and make the subsequent learning within the neurons much easier. 

\begin{table}[!]
  \centering
  \caption{Effectiveness of EleAttG on the hidden state \textbf{h$_{t-1}$} and the input vector \textbf{x$_t$} in terms of accuracy (\%) on the NTU dataset.}
    \begin{tabular}{ccc}
    \toprule
    Method & CS    & CV \\
    \midrule
    Baseline-GRU & 75.2  & 81.5 \\
    EleAtt-GRU(\textbf{h$_{t-1}$}) & 79.3  & 86.1 \\
    EleAtt-GRU(\textbf{x$_t$}) & \textbf{79.8}  & \textbf{87.1} \\
    EleAtt-GRU(\textbf{h$_{t-1}$} \& \textbf{x$_t$}) & 79.0    & 85.4 \\
    \bottomrule
    \end{tabular}
  \label{tab:EleAttG}
\end{table}

\setlength{\tabcolsep}{4pt}
\begin{table*}[th]
	\centering
	\caption{Performance comparisons about relaxing the constraint to EleAttG on the NTU dataset in terms of accuracy (\%).}
	\label{tab:constrain}
	\begin{tabular}{cccccc}
		\toprule
		Protocols           & Method              & Baseline-GRU & EleAttG-$1^{st}$ & EleAttG-$2^{nd}$ & EleAttG-$3^{rd}$ \\
		\midrule
		\multirow{2}{*}{CS} &  w/ constraint  & 75.2          & 75.0      & 72.7      & 72.0      \\
		&  wo/ constrain & 75.2          & \bf{78.7}      & \bf{77.3}      & \bf{76.4}      \\
		\midrule
		\multirow{2}{*}{CV} &  w/ constraint  & 81.5          & 83.7      & 79.1      & 78.8      \\
		&  wo/ constrain & 81.5          & \bf{84.9}      & \bf{83.5}      & \bf{82.5}     \\
		\bottomrule
	\end{tabular}
\end{table*}
\textbf{Applying EleAttG to x$_t$ or h$_{t-1}$?} We experimentally show the effectiveness of EleAttG when it operates on the hidden state \textbf{h$_{t-1}$} and the input vector \textbf{x$_t$} respectively in Table \ref{tab:EleAttG}. Note that for a time slot $t$, besides the input vector \textbf{x$_t$}, the hidden state of the last time slot \textbf{h$_{t-1}$} is another input vector.

``Baseline-GRU" denotes the baseline model that is built by three GRU layers without EleAttG. 

``EleAtt-GRU(\textbf{h$_{t-1}$})" denotes the model that is built by three GRU layers with EleAttG, where the EleAttG is used to modify the hidden state \textbf{h$_{t-1}$}, which is also the input vector of the current time slot.

``EleAtt-GRU(\textbf{x$_t$})" denotes the model that is built by three GRU layers with EleAttG, where the EleAttG is used to modify the input vector \textbf{x$_t$}. 

``EleAtt-GRU(\textbf{h$_{t-1}$} $\&$ \textbf{x$_t$})" denotes the model that is built by three GRU layers with two EleAttGs, which are used to modify the hidden state \textbf{h$_{t-1}$} and the input vector \textbf{x$_t$}, respectively.

From Table \ref{tab:EleAttG}, we observe that accuracy improves when the EleAttG is used to the hidden state \textbf{h$_{t-1}$} or the input vector \textbf{x$_t$}, respectively. ``EleAtt-GRU(\textbf{x$_t$})" outperforms ``EleAtt-GRU(\textbf{h$_{t-1}$})" by 0.5\% and 1.0\% for the CS and CV settings, which demonstrates that applying the EleAttG to the input vector \textbf{x$_t$} is more effective. However, when applying the EleAttGs to both the hidden state \textbf{h$_{t-1}$} and the input vector \textbf{x$_t$} simultaneously, performance does not improve further and even becomes poorer in comparison with the cases when applying the EleAttG to the hidden state \textbf{h$_{t-1}$} or the input vector \textbf{x$_t$}, respectively. One potential reason is that the number of parameters increases when applying two EleAttGs simultaneously in comparison with the model using only one EleAttG, which makes it harder to optimize the model. In addition, the physical meaning of applying modulation on \textbf{x$_t$} is more clear than that on \textbf{h$_{t-1}$}. On the other hand, whenever suitable attention is applied to the input \textbf{x$_t$}, the output \textbf{h$_{t}$} may already be the attended feature for the next time slot $t+1$.

\textbf{Relaxing the sum-to-1 constraint on \EleAttG responses.} Unlike other works \cite{xu2015show,song2017end,liu2017global}, we do not use Softmax, which enforces the sum of attention responses to be 1, as the activation function of EleAttG. On the contrary, we use the Sigmoid activation function to avoid introducing  mutual influence of elements. If the sum-to-1 constraint is not relaxed, the attention response of the $k^{th}$ element will be affected by the changes of other elements' response values even when the levels of importance of this element are the same over consecutive time slots. Especially for a sequence, the constraint could break the continuity of the features/inputs after applying attention.

We show the experimental comparisons between the cases with the sum-to-1 constraint ($w/ constraint$) by using Softmax, and our case without such constraint ($wo/ constraint$) by using Sigmoid in Table \ref{tab:constrain}. ``EleAttG-$n^{th}$" denotes that the $n^{th}$ GRU layer uses the GRU with EleAttG while the other layers still use the original GRU. ``Baseline-GRU" denotes the baseline scheme with three GRU layers. We can see that $wo/ constraint$ always performs better than that with constraint $w/ constraint$. Specially, adding EleAttG with constraint on the second or the third layer even decreases the accuracy by about 2.4-3.2\% in comparison with the baselines. 

\textbf{Different from the input gate of LSTM model.}  An input gate is designed to control the contribution of the current input to memory versus the contribution of the historical information controlled by a forget gate. It uses a scalar (rather than a vector) to control the contribution of the current input vector. The proposed element-wise attention gate (EleAttG) applies element-wise adaptive modulation to the input vector to achieve element-wise attention before further processing in the RNN neurons. EleAttG is designed to control the contribution of {\emph{each element}} of the current input to the RNN by suppressing the amplitudes of the unimportant elements while preserving the amplitudes of the important elements.

We can modify our proposed EleAttG to produce a scalar to represent the importance of the current input vector, where all elements of the input vector share the same importance value. We denote the modified EleAttG as EleAttG$^1$. The scheme with EleAttG is denoted as ``EleAtt-GRU" and the scheme with EleAttG$^1$ is denoted as ``EleAtt$^1$-GRU". Baseline-GRU denotes the schemes without EleAttGs. We compare the ``Baseline-GRU", ``EleAtt$^1$-GRU'', and  ``EleAtt-GRU" in Table \ref{tab:gate}. We observe that both ``EleAtt-GRU" and ``EleAtt$^1$-GRU" are superior to the ``Basleine-GRU". Our proposed ``EleAtt-GRU" outperforms ``EleAtt$^1$-GRU" by 3.0\% and 3.2\% for the CS and CV settings, respectively.

\setlength{\tabcolsep}{6pt}
\begin{table}[!]
  \centering
  \caption{Comparisons about ``EleAtt-GRU" and ``EleAtt$^1$-GRU" on the NTU dataset in terms of accuracy (\%).  ``EleAtt$^1$-GRU" denotes all elements of the input vector share the same importance level.}
    \begin{tabular}{ccc}
    \toprule
    Method & CS  & CV \\
    \midrule
    Baselin-GRU & 75.2 & 81.5 \\
    EleAtt$^1$-GRU & 76.8 & 83.9 \\
    EleAtt-GRU & \textbf{79.8} & \textbf{87.1} \\
    \bottomrule
    \end{tabular}
  \label{tab:gate}
\end{table}

\setlength{\tabcolsep}{6pt}
\begin{table}[t]
	\centering
	\caption{Effect of the number of parameters  on the NTU dataset.}
	\begin{tabular}{cccc}
		\toprule
		Scheme & \# Parameters & CS    & CV \\
		\midrule
		2-GRU(100) & 0.14M & 75.5  & 81.4  \\
		2-GRU(128) & 0.21M & 75.8 & 81.7 \\
		3-GRU(100) & 0.20M & 75.2  & 81.5  \\
		3-GRU(128) & 0.31M & 76.5  & 81.3  \\
		2-EleAtt-GRU(100) & 0.20M & 78.6  & 85.5 \\
		3-EleAtt-GRU(100) & 0.28M & 79.8  & 87.1  \\
		\bottomrule
	\end{tabular}
	\label{tab:params}
\end{table}

\textbf{Number of parameters versus performance.} For an RNN block, the adding of an EleAttG increases the number of parameters. One may wonder whether the performance is increased by {\EleAttGn} or just the additional parameters. We analyze the influence of parameters as follows.

Taking a GRU block consisting of $N$ neurons with the input dimension of $D$ as an example, the numbers of parameters for the original GRU block and the proposed EleAttG-GRU block are $3N(D+N+1)$, and $3N(D+N+1) +D(D+N+1)$, respectively. We calculate the computational complexity by counting the number of floating-point operations (FLOPs) including all multiplication and addition operations. At a time slot, adding attention to the layer as in \ref{equ:agru} and \ref{equ:updatedx} takes $D(D+N+1)$ multiplication operations and $D(D+N)$ addition operations. Then the complexity increases from $N(6D+6N+5)$ to $N(6D+6N+5)+D(2D+2N+1)$, which is approximately proportional to the number of parameters. 

Table \ref{tab:params} shows the effect of the number of parameters under different experimental settings on the NTU dataset. Note that ``$m$-GRU($n$)" denotes the baseline scheme which is built by $m$ GRU blocks (layers) with each layer composed of $n$ neurons. ``$m$-EleAtt-GRU(100)" denotes our scheme which includes $m$ EleAtt-GRU layers with each layer composed of 100 neurons.  We can see that the performance increases only a little when more neurons (``2-GRU(128)") or more layers (``3-GRU(100)") are used in comparison with the baseline ``2-GRU(100)". In contrast, our scheme ``2-EleAtt-GRU(100)", achieves significant gains of 3.1-4.1\% in comparison with ``2-GRU(100)". Similar observations are made in three-layer cases. With similar numbers of parameters, adding \EleAttG is much more effective than increasing the number of neurons or the number of layers. It demonstrates that \EleAttG significantly boosts performance.

\section{Conclusions}

In this paper, we propose a simple yet effective \EleAttG to empower the neurons in recurrent neural networks to have the attentiveness capability. It can explore the varying importance of different element of the input. Experiments show that our proposed EleAttG can be used in any RNN structures ({\it{e.g}} standard RNN, LSTM and GRU), any layers of the multi-layer RNN networks, and different types of input signals ({\it{e.g}} skeleton data, CNN features, and raw image pixels). Abundant experiments show that the proposed EleAttG boosts the performance significantly. We expect that, as a fundamental unit, the proposed EleAttG will be effective for improving many RNN-based learning tasks.

\section*{Acknowledgment}
This work is supported by National Natural Science Foundation of China under Grant 61751308 and Grant 61773311, and National Key Research and Development Program of China under Grant 2016YFB1001004.


%

%



\bibliographystyle{abbrv}
\bibliography{egbib}

\begin{thebibliography}{10}

\bibitem{al2016theano}
R.~Al-Rfou, G.~Alain, A.~Almahairi, C.~Angermueller, D.~Bahdanau, N.~Ballas,
  F.~Bastien, J.~Bayer, A.~Belikov, A.~Belopolsky, et~al.
\newblock Theano: A python framework for fast computation of mathematical
  expressions.
\newblock {\em arXiv}, 2016.

\bibitem{arjovsky2016unitary}
M.~Arjovsky, A.~Shah, and Y.~Bengio.
\newblock Unitary evolution recurrent neural networks.
\newblock In {\em ICML}, 2016.

\bibitem{campos2018skip}
V.~Campos~Camunez, B.~Jou, X.~Gir{\'o}~Nieto, J.~Torres~Vi{\~n}als, and S.-F.
  Chang.
\newblock Skip rnn: learning to skip state updates in recurrent neural
  networks.
\newblock In {\em ICLR}, 2018.

\bibitem{cho14}
K.~Cho, B.~van Merri{\"{e}}nboer, {\c C}.~G{\"{u}}l{\c c}ehre, D.~Bahdanau,
  F.~Bougares, H.~Schwenk, and Y.~Bengio.
\newblock Learning phrase representations using rnn encoder--decoder for
  statistical machine translation.
\newblock In {\em EMNLP}, Oct. 2014.

\bibitem{ResNet50Model}
F.~Chollet.
\newblock Resnet50 model.
\newblock
  \url{https://github.com/fchollet/deep-learning-models/releases/download/v0.2/resnet50_weights_tf_dim_ordering_tf_kernels.h5}.

\bibitem{chollet2015keras}
F.~Chollet.
\newblock Keras.
\newblock \url{https://github.com/fchollet/keras}, 2015.

\bibitem{chung2014empirical}
J.~Chung, C.~Gulcehre, K.~Cho, and Y.~Bengio.
\newblock Empirical evaluation of gated recurrent neural networks on sequence
  modeling.
\newblock In {\em NIPSW}, 2014.

\bibitem{cooijmans2016recurrent}
T.~Cooijmans, N.~Ballas, C.~Laurent, {\c{C}}.~G{\"u}l{\c{c}}ehre, and
  A.~Courville.
\newblock Recurrent batch normalization.
\newblock {\em arXiv}, 2016.

\bibitem{de2016skeleton}
Q.~De~Smedt, H.~Wannous, and J.-P. Vandeborre.
\newblock Skeleton-based dynamic hand gesture recognition.
\newblock In {\em CVPRW}, 2016.

\bibitem{diba2017deeptemporal}
A.~Diba, V.~Sharma, and L.~Van~Gool.
\newblock Deep temporal linear encoding networks.
\newblock In {\em CVPR}, 2017.

\bibitem{donahue2015long}
J.~Donahue, L.~Anne~Hendricks, S.~Guadarrama, M.~Rohrbach, S.~Venugopalan,
  K.~Saenko, and T.~Darrell.
\newblock Long-term recurrent convolutional networks for visual recognition and
  description.
\newblock In {\em CVPR}, 2015.

\bibitem{du2016representation}
Y.~Du, Y.~Fu, and L.~Wang.
\newblock Representation learning of temporal dynamics for skeleton-based
  action recognition.
\newblock {\em TIP}, 25(7):3010--3022, 2016.

\bibitem{du2015hierarchical}
Y.~Du, W.~Wang, and L.~Wang.
\newblock Hierarchical recurrent neural network for skeleton based action
  recognition.
\newblock In {\em CVPR}, 2015.

\bibitem{evangelidis2014skeletal}
G.~Evangelidis, G.~Singh, and R.~Horaud.
\newblock Skeletal quads: Human action recognition using joint quadruples.
\newblock In {\em ICPR}, 2014.

\bibitem{gers1999learning}
F.~A. Gers, J.~Schmidhuber, and F.~Cummins.
\newblock Learning to forget: Continual prediction with lstm.
\newblock 1999.

\bibitem{gers2002learning}
F.~A. Gers, N.~N. Schraudolph, and J.~Schmidhuber.
\newblock Learning precise timing with lstm recurrent networks.
\newblock {\em JMLR}, 3(Aug):115--143, 2002.

\bibitem{he2016deep}
K.~He, X.~Zhang, S.~Ren, and J.~Sun.
\newblock Deep residual learning for image recognition.
\newblock In {\em CVPR}, 2016.

\bibitem{hochreiter1997long}
S.~Hochreiter and J.~Schmidhuber.
\newblock Long short-term memory.
\newblock {\em Neural computation}, 9(8):1735--1780, 1997.

\bibitem{hu2018squeeze}
J.~Hu, L.~Shen, and G.~Sun.
\newblock Squeeze-and-excitation networks.
\newblock In {\em CVPR}, 2018.

\bibitem{hu2015jointly}
J.-F. Hu, W.-S. Zheng, J.~Lai, and J.~Zhang.
\newblock Jointly learning heterogeneous features for rgb-d activity
  recognition.
\newblock In {\em CVPR}, 2015.

\bibitem{hu2016real}
J.-F. Hu, W.-S. Zheng, L.~Ma, G.~Wang, and J.~Lai.
\newblock Real-time rgb-d activity prediction by soft regression.
\newblock In {\em ECCV}, 2016.

\bibitem{jhuang2013towards}
H.~Jhuang, J.~Gall, S.~Zuffi, C.~Schmid, and M.~J. Black.
\newblock Towards understanding action recognition.
\newblock In {\em ICCV}, 2013.

\bibitem{jozefowicz2015empirical}
R.~Jozefowicz, W.~Zaremba, and I.~Sutskever.
\newblock An empirical exploration of recurrent network architectures.
\newblock In {\em ICML}, 2015.

\bibitem{ke2017new}
Q.~Ke, M.~Bennamoun, S.~An, F.~Sohel, and F.~Boussaid.
\newblock A new representation of skeleton sequences for 3d action recognition.
\newblock In {\em CVPR}. IEEE, 2017.

\bibitem{kingma2014adam}
D.~P. Kingma and J.~Ba.
\newblock Adam: A method for stochastic optimization.
\newblock {\em arXiv}, 2014.

\bibitem{kuehne2011hmdb}
H.~Kuehne, H.~Jhuang, E.~Garrote, T.~Poggio, and T.~Serre.
\newblock Hmdb: a large video database for human motion recognition.
\newblock In {\em ICCV}, 2011.

\bibitem{kusupati2018fastgrnn}
A.~Kusupati, M.~Singh, K.~Bhatia, A.~Kumar, P.~Jain, and M.~Varma.
\newblock Fastgrnn: A fast, accurate, stable and tiny kilobyte sized gated
  recurrent neural network.
\newblock In {\em NIPS}, 2018.

\bibitem{le2015simple}
Q.~V. Le, N.~Jaitly, and G.~E. Hinton.
\newblock A simple way to initialize recurrent networks of rectified linear
  units.
\newblock {\em arXiv}, 2015.

\bibitem{lecun1998gradient}
Y.~LeCun, L.~Bottou, Y.~Bengio, and P.~Haffner.
\newblock Gradient-based learning applied to document recognition.
\newblock {\em Proceedings of the IEEE}, 86(11):2278--2324, 1998.

\bibitem{li2017attentive}
J.~Li, Y.~Wei, X.~Liang, J.~Dong, T.~Xu, J.~Feng, and S.~Yan.
\newblock Attentive contexts for object detection.
\newblock {\em TMM}, 19(5):944--954, 2017.

\bibitem{li2018independently}
S.~Li, W.~Li, C.~Cook, C.~Zhu, and Y.~Gao.
\newblock Independently recurrent neural network (indrnn): Building a longer
  and deeper rnn.
\newblock In {\em CVPR}, 2018.

\bibitem{li2017adaptive}
W.~Li, L.~Wen, M.-C. Chang, S.~N. Lim, and S.~Lyu.
\newblock Adaptive rnn tree for large-scale human action recognition.
\newblock In {\em CVPR}, 2017.

\bibitem{liu2016spatio}
J.~Liu, A.~Shahroudy, D.~Xu, and G.~Wang.
\newblock Spatio-temporal lstm with trust gates for 3d human action
  recognition.
\newblock In {\em ECCV}, 2016.

\bibitem{liu2017global}
J.~Liu, G.~Wang, P.~Hu, L.-Y. Duan, and A.~C. Kot.
\newblock Global context-aware attention lstm networks for 3d action
  recognition.
\newblock In {\em CVPR}, 2017.

\bibitem{luong2015effective}
T.~Luong, H.~Pham, and C.~D. Manning.
\newblock Effective approaches to attention-based neural machine translation.
\newblock In {\em EMNLP}, 2015.

\bibitem{nunez2018convolutional}
J.~C. N{\'u}{\~n}ez, R.~Cabido, J.~J. Pantrigo, A.~S. Montemayor, and J.~F.
  V{\'e}lez.
\newblock Convolutional neural networks and long short-term memory for
  skeleton-based human activity and hand gesture recognition.
\newblock {\em PR}, 76:80--94, 2018.

\bibitem{LSTMblog}
C.~Olah.
\newblock Lstm.
\newblock \url{http://colah.github.io/posts/2015-08-Understanding-LSTMs/},
  2015.

\bibitem{Shahroudy_2016_CVPR}
A.~Shahroudy, J.~Liu, T.-T. Ng, and G.~Wang.
\newblock Ntu rgb+d: A large scale dataset for 3d human activity analysis.
\newblock In {\em CVPR}, 2016.

\bibitem{sharma2015actionattention}
S.~Sharma, R.~Kiros, and R.~Salakhutdinov.
\newblock Action recognition using visual attention.
\newblock {\em arXiv}, 2015.

\bibitem{si2018skeleton}
C.~Si, Y.~Jing, W.~Wang, L.~Wang, and T.~Tan.
\newblock Skeleton-based action recognition with spatial reasoning and temporal
  stack learning.
\newblock {\em ECCV}, 2018.

\bibitem{simonyan2014two}
K.~Simonyan and A.~Zisserman.
\newblock Two-stream convolutional networks for action recognition in videos.
\newblock In {\em NIPS}, 2014.

\bibitem{song2017end}
S.~Song, C.~Lan, J.~Xing, W.~Zeng, and J.~Liu.
\newblock An end-to-end spatio-temporal attention model for human action
  recognition from skeleton data.
\newblock In {\em AAAI}, 2017.

\bibitem{song2018spatio}
S.~Song, C.~Lan, J.~Xing, W.~Zeng, and J.~Liu.
\newblock Spatio-temporal attention-based lstm networks for 3d action
  recognition and detection.
\newblock {\em TIP}, 27(7):3459--3471, 2018.

\bibitem{srivastava2014dropout}
N.~Srivastava, G.~Hinton, A.~Krizhevsky, I.~Sutskever, and R.~Salakhutdinov.
\newblock Dropout: A simple way to prevent neural networks from overfitting.
\newblock {\em JMLR}, 15(1):1929--1958, 2014.

\bibitem{sutskever2014sequence}
I.~Sutskever, O.~Vinyals, and Q.~V. Le.
\newblock Sequence to sequence learning with neural networks.
\newblock In {\em NIPS}, 2014.

\bibitem{vaswani2017attention}
A.~Vaswani, N.~Shazeer, N.~Parmar, J.~Uszkoreit, L.~Jones, A.~N. Gomez,
  {\L}.~Kaiser, and I.~Polosukhin.
\newblock Attention is all you need.
\newblock In {\em NIPS}, 2017.

\bibitem{veeriah2015differential}
V.~Veeriah, N.~Zhuang, and G.-J. Qi.
\newblock Differential recurrent neural networks for action recognition.
\newblock In {\em ICCV}, 2015.

\bibitem{vemulapalli2014human}
R.~Vemulapalli, F.~Arrate, and R.~Chellappa.
\newblock Human action recognition by representing 3d skeletons as points in a
  lie group.
\newblock In {\em CVPR}, 2014.

\bibitem{vinyals2015show}
O.~Vinyals, A.~Toshev, S.~Bengio, and D.~Erhan.
\newblock Show and tell: A neural image caption generator.
\newblock In {\em CVPR}, 2015.

\bibitem{wang2018dividing}
D.~Wang, W.~Ouyang, W.~Li, and D.~Xu.
\newblock Dividing and aggregating network for multi-view action recognition.
\newblock In {\em ECCV}, 2018.

\bibitem{wang2013action}
H.~Wang and C.~Schmid.
\newblock Action recognition with improved trajectories.
\newblock In {\em ICCV}, 2013.

\bibitem{wang2013learning}
J.~Wang, Z.~Liu, Y.~Wu, and J.~Yuan.
\newblock Learning actionlet ensemble for 3d human action recognition.
\newblock {\em TPAMI}, 36(5):914--927, 2013.

\bibitem{wang2014cross}
J.~Wang, X.~Nie, Y.~Xia, Y.~Wu, and S.-C. Zhu.
\newblock Cross-view action modeling, learning and recognition.
\newblock In {\em CVPR}, 2014.

\bibitem{wang2016temporal}
L.~Wang, Y.~Xiong, Z.~Wang, Y.~Qiao, D.~Lin, X.~Tang, and L.~Van~Gool.
\newblock Temporal segment networks: Towards good practices for deep action
  recognition.
\newblock In {\em ECCV}, 2016.

\bibitem{wang2016graph}
P.~Wang, C.~Yuan, W.~Hu, B.~Li, and Y.~Zhang.
\newblock Graph based skeleton motion representation and similarity measurement
  for action recognition.
\newblock In {\em ECCV}, 2016.

\bibitem{wang2016hierarchical}
Y.~Wang, S.~Wang, J.~Tang, N.~O'Hare, Y.~Chang, and B.~Li.
\newblock Hierarchical attention network for action recognition in videos.
\newblock {\em arXiv}, 2016.

\bibitem{weng2018deformable}
J.~Weng, M.~Liu, X.~Jiang, and J.~Yuan.
\newblock Deformable pose traversal convolution for 3d action and gesture
  recognition.
\newblock In {\em ECCV}, 2018.

\bibitem{weng2017spatio}
J.~Weng, C.~Weng, and J.~Yuan.
\newblock Spatio-temporal naive-bayes nearest-neighbor (st-nbnn) for
  skeleton-based action recognition.
\newblock In {\em CVPR}, 2017.

\bibitem{woo2018cbam}
S.~Woo, J.~Park, J.-Y. Lee, and I.~So~Kweon.
\newblock Cbam: Convolutional block attention module.
\newblock In {\em ECCV}, 2018.

\bibitem{xia2012view}
L.~Xia, C.-C. Chen, and J.~Aggarwal.
\newblock View invariant human action recognition using histograms of 3d
  joints.
\newblock In {\em CVPRW}, 2012.

\bibitem{TSNModel}
Y.~Xiong.
\newblock {TSN} model.
\newblock \url{https://github.com/yjxiong/temporal-segment-networks}, 2016.

\bibitem{xu2015show}
K.~Xu, J.~Ba, R.~Kiros, K.~Cho, A.~Courville, R.~Salakhudinov, R.~Zemel, and
  Y.~Bengio.
\newblock Show, attend and tell: Neural image caption generation with visual
  attention.
\newblock In {\em ICML}, 2015.

\bibitem{yue2015beyond}
J.~Yue-Hei~Ng, M.~Hausknecht, S.~Vijayanarasimhan, O.~Vinyals, R.~Monga, and
  G.~Toderici.
\newblock Beyond short snippets: Deep networks for video classification.
\newblock In {\em CVPR}, 2015.

\bibitem{zhang2017view}
P.~Zhang, C.~Lan, J.~Xing, W.~Zeng, J.~Xue, and N.~Zheng.
\newblock View adaptive recurrent neural networks for high performance human
  action recognition from skeleton data.
\newblock In {\em ICCV}, 2017.

\bibitem{zhang2019view}
P.~Zhang, C.~Lan, J.~Xing, W.~Zeng, J.~Xue, and N.~Zheng.
\newblock View adaptive neural networks for high performance skeleton-based
  human action recognition.
\newblock {\em TPAMI}, 2019.

\bibitem{zhang2019sr}
P.~Zhang, W.~Ouyang, P.~Zhang, J.~Xue, and N.~Zheng.
\newblock Sr-lstm: State refinement for lstm towards pedestrian trajectory
  prediction.
\newblock In {\em CVPR}, 2019.

\bibitem{zhang2018adding}
P.~Zhang, J.~Xue, C.~Lan, W.~Zeng, Z.~Gao, and N.~Zheng.
\newblock Adding attentiveness to the neurons in recurrent neural networks.
\newblock In {\em ECCV}, 2018.

\bibitem{zhang2016architectural}
S.~Zhang, Y.~Wu, T.~Che, Z.~Lin, R.~Memisevic, R.~R. Salakhutdinov, and
  Y.~Bengio.
\newblock Architectural complexity measures of recurrent neural networks.
\newblock In {\em NIPS}, 2016.

\bibitem{zhao2018hsa}
B.~Zhao, X.~Li, and X.~Lu.
\newblock Hsa-rnn: Hierarchical structure-adaptive rnn for video summarization.
\newblock In {\em CVPR}, 2018.

\bibitem{zhu2016co}
W.~Zhu, C.~Lan, J.~Xing, W.~Zeng, Y.~Li, L.~Shen, X.~Xie, et~al.
\newblock Co-occurrence feature learning for skeleton based action recognition
  using regularized deep lstm networks.
\newblock In {\em AAAI}, 2016.

\end{thebibliography}
\ifCLASSOPTIONcaptionsoff
  \newpage
\fi

\end{document}